\documentclass[11pt]{article}

\usepackage[final]{acl}

\usepackage{times}
\usepackage{latexsym}

\usepackage[T1]{fontenc}

\usepackage[utf8]{inputenc}

\usepackage{microtype}

\usepackage{inconsolata}

\usepackage{graphicx}
\usepackage{amsmath}
\usepackage{amssymb}
\usepackage{booktabs}
\usepackage{multirow}
\usepackage{array}
\usepackage{listings}
\usepackage{xcolor}
\usepackage{tabularx}
\usepackage[table]{xcolor}

\colorlet{punct}{red!60!black}
\definecolor{background}{HTML}{EEEEEE}
\definecolor{delim}{RGB}{20,105,176}
\colorlet{numb}{magenta!60!black}

\lstdefinelanguage{json}{
    basicstyle=\normalfont\ttfamily\footnotesize,
    stepnumber=1,
    numbersep=8pt,
    showstringspaces=false,
    breaklines=true,
    frame=lines,
    literate=
     *{0}{{{\color{numb}0}}}{1}
      {1}{{{\color{numb}1}}}{1}
      {2}{{{\color{numb}2}}}{1}
      {3}{{{\color{numb}3}}}{1}
      {4}{{{\color{numb}4}}}{1}
      {5}{{{\color{numb}5}}}{1}
      {6}{{{\color{numb}6}}}{1}
      {7}{{{\color{numb}7}}}{1}
      {8}{{{\color{numb}8}}}{1}
      {9}{{{\color{numb}9}}}{1}
      {:}{{{\color{punct}{:}}}}{1}
      {,}{{{\color{punct}{,}}}}{1}
      {\{}{{{\color{delim}{\{}}}}{1}
      {\}}{{{\color{delim}{\}}}}}{1}
      {[}{{{\color{delim}{[}}}}{1}
      {]}{{{\color{delim}{]}}}}{1},
}

\newcommand\Benchmark{NaturalGAIA}
\newcommand\Agent{LightManus}
\newcommand\AndroidA{Jarvis}

\title{\Benchmark: A Verifiable Benchmark and Hierarchical Framework for Long-Horizon GUI Tasks}

\author{
    Zihan Zheng$^{1}$\footnotemark[1], \quad
    Tianle Cui$^{1}$\footnotemark[1], \quad
    Taoran Wang$^{1}$, \quad
    Fengtao Wang$^{1}$, \\
    \textbf{Jiahui Pan$^{1}$}, \quad
    \textbf{Lewei He$^{1}$\footnotemark[2]}, \quad
    \textbf{Qianglong Chen$^{2}$}\footnotemark[2] \\
    $^1$South China Normal University \quad $^2$Zhejiang University \\
    \texttt{zhengzihan994@m.scnu.edu.cn} \quad
    \texttt{helewei@m.scnu.edu.cn} \quad
    \texttt{chenqianglong@zju.edu.cn}
}

\begin{document}
\maketitle

{
  \renewcommand{\thefootnote}%
  {\fnsymbol{footnote}}
  \footnotetext[1]{Equal contribution.}
  \footnotetext[2]{Corresponding Authors.}
}

\begin{abstract}
Despite significant advances in LLM-driven GUI agents, the field remains constrained by the challenge of reconciling high-fidelity realism with verifiable evaluation accuracy. To address this, we introduce \Benchmark, a verifiable evaluation dataset grounded in real-world human GUI interaction intents. By decoupling logical causal pathways from linguistic narratives, it rigorously simulates natural human intent, characterized by cognitive non-linearity and contextual dependencies. Furthermore, we propose \Agent-\AndroidA, a hierarchical collaborative framework where \Agent~manages dynamic topological planning and context evolution, while \AndroidA~ensures execution precision via hybrid visual-structural perception. Experiments demonstrate that our approach achieves a Weighted Pathway Success Rate of 45.6\%, significantly outperforming the state-of-the-art baseline (21.1\%), while reducing token consumption by 75\% and execution time by 76\%. These results validate the efficacy of the macro-planning and micro-execution paradigm in handling complex naturalized tasks. Our code is publicly available at: https://github.com/KeLes-Coding/NatureGAIA.
\end{abstract}

\section{Introduction}
\begin{figure*}[t]
    \centering
    \includegraphics[width=1\linewidth]{source/pic/main_v1_2512125.pdf}
    \caption{Overview of the \textbf{\Agent-\AndroidA}~framework executing a task from \textbf{\Benchmark}. \textbf{Task Parsing}: The parser decomposes abstract user intents into a structured Task Topology composed of atomic tasks. \textbf{Workflow Management}: The manager dynamically schedules tasks across heterogeneous agents, employing Context Evolution to bridge information gaps between steps. \textbf{Execution \& Evaluation}: \AndroidA~execute actions via hybrid perception, assessed by a hierarchical framework for success rates and error attribution.}
    \label{fig:main_pic}
\end{figure*}
Although LLM-driven GUI agents have advanced significantly \citep{deepseekai2025deepseekr1incentivizingreasoningcapability, openai2024openaio1card, qwen2025qwen25technicalreport}, current research faces an "Evaluation-Realism Dilemma" \citep{10.1145/3711896.3736570, gera-etal-2025-justrank}. Realistic benchmarks like OSWorld \citep{OSWorld} and RealWebAssist \citep{ye2025realwebassistbenchmarklonghorizonweb} lack deterministic ground truth due to reliance on unstable MLLM judges or manual verification, preventing accurate measurement of the "reasoning-execution gap" \citep{dong2025saythinganotherdiagnosing, liu-etal-2025-care}. Conversely, traditional static benchmarks \citep{deng2023mind2webgeneralistagentweb, he-etal-2024-webvoyager, zhou2023webarena} use simplified, decontextualized instructions that fail to capture the cognitive non-linearity of human intent \citep{ye2025realwebassistbenchmarklonghorizonweb, zhang2025agenticcontextengineeringevolving}. However, these fail to capture the cognitive non-linearity of human intent \citep{dong2025saythinganotherdiagnosing, wei-etal-2025-plangenllms}, leading to overestimated success rates that do not reflect robustness in unstructured environments \citep{dai2025scubasalesforcecomputeruse, kartik2025agentcompassreliableevaluationagentic, he2025efficientagenttrainingcomputer}.

Simultaneously, existing architectures face stability issues in long-range, noisy workflows. Mainstream end-to-end vision models \citep{wang2025uitars2technicalreportadvancing, niu2024screenagent} often prioritize visual "shortcut learning" over generalizable procedural knowledge \citep{yang2023appagent, cheng-etal-2024-seeclick}. This monolithic design suffers from a dual bottleneck: the lack of dynamic context management causes semantic drift in long sequences \citep{zhang2025agenticcontextengineeringevolving, cai2025flexcontinuousagentevolution}, while pure visual perception is prone to "coordinate hallucinations" in resource-constrained environments, hindering fine-grained precision \citep{yang2025ferretuilitelessonsbuilding, yin2025hypernavhybridperceptionobjectoriented}. Therefore, balancing macroscopic planning coherence with microscopic execution precision \citep{he2024pcagent} remains a core unresolved challenge.

To address the limitations of existing evaluation methods, we introduce \Benchmark, a dynamic dataset currently comprising 276 constructed tasks. Its core innovation lies in separating underlying logical structure from linguistic representation via knowledge-driven causal paths. By integrating a naturalized narrative layer with a multi-level evaluation framework, \Benchmark~rigorously assesses agent performance and isolates failure sources, ranging from intent parsing to execution planning. This design ensures a focused evaluation of core capabilities, including implicit intent understanding, long-horizon planning, and complex tool usage.

To overcome the execution challenges inherent in long-horizon tasks, we propose \Agent, a hierarchical collaborative framework. Adopting a "macro-planning-micro-execution" paradigm, the system operates on two levels: at the macro level, a workflow manager orchestrates task topologies and manages context evolution; at the micro level, \AndroidA~serves as an efficient execution kernel, achieving precise, coordinate-independent atomic actions via hybrid perception and chain-of-thought reasoning. This architecture effectively bridges the gap between high-level semantic dependencies and low-level operational barriers.

We evaluate our approach using \Benchmark. Experimental results demonstrate that the \Agent-\AndroidA~framework (driven by Claude-Sonnet-4.5) achieves a Weighted Pathway Success Rate (WPSR) of 45.6\%, significantly outperforming mainstream baselines such as PC-Agent and Mobile-Agent-e, which score 13.1\% and 21.1\% respectively. Furthermore, our architecture improves operational efficiency by reducing token consumption by approximately 75\% compared to existing GUI agents. These findings highlight the effectiveness of the hierarchical architecture in managing long-term context and precise execution, while also revealing the remaining challenges in handling highly ambiguous naturalized narratives.

The main contributions of this paper can be summarized as follows:
\begin{itemize}
    \item [1)] Propose a new benchmark: \Benchmark, which separates logic and language by causal paths and narrative layers to evaluate the task performance of intelligent agents.
    \item [2)] We design the \Agent-\AndroidA~layered architecture, integrating dynamic workflow scheduling and a high-precision GUI execution kernel.
    \item [3)] We complete a comprehensive empirical evaluation, revealing the key shortcomings of current agents in complex tasks.
\end{itemize}

\section{Related Work}
GUI agent research is transitioning from static web parsing \cite{zhou2023webarena, deng2023mind2webgeneralistagentweb} to dynamic OS control \cite{OSWorld, chai2025a3androidagentarena, rawles2024androidworld}, yet reconciling ecological validity with reproducibility remains a core challenge \cite{riddell-etal-2024-quantifying, fang2024bidboundaryinteriordecodingunsupervised, lin2025cuarewardbenchbenchmarkevaluatingreward, zheng-etal-2025-planningarena}. Specifically, realistic benchmarks \cite{ye2025realwebassistbenchmarklonghorizonweb, xu2024agenttrek} often lack deterministic verification \cite{drouin2024workarenacapablewebagents}, relying on unstable or costly evaluations \cite{dai2025scubasalesforcecomputeruse, lù2025agentrewardbenchevaluatingautomaticevaluations}. Conversely, deterministic approaches \cite{sun2025scienceboardevaluatingmultimodalautonomous, valmeekam2023planbenchextensiblebenchmarkevaluating, pan2025spiderscalablephysicsinformeddexterous} tend to simplify tasks, failing to assess open-world planning \cite{dong2025saythinganotherdiagnosing, kartik2025agentcompassreliableevaluationagentic} or safety \cite{andriushchenko2025agentharmbenchmarkmeasuringharmfulness}. regarding architecture, end-to-end LMMs \cite{wang2025uitars2technicalreportadvancing, niu2024screenagent, andreux2025surfer2generationcrossplatform} offer strong generalization but face computational constraints on mobile devices \cite{yang2025ferretuilitelessonsbuilding, yin2025hypernavhybridperceptionobjectoriented, zhang-etal-2025-agentcpm}. Meanwhile, modular frameworks \cite{he2024pcagent, li2025hiplanhierarchicalplanningllmbased, hong2024metagptmetaprogrammingmultiagent, shang2025agentsquareautomaticllmagent} improve interpretability but struggle with context drift and cross-modal alignment in long sequences \cite{huang2025guikvefficientguiagents, mao2025liftimprovinglongcontext, zhang2025agenticcontextengineeringevolving, cai2025flexcontinuousagentevolution}. Moreover, integrating symbolic logic with neural planning \cite{choi2025nesycneurosymboliccontinuallearner, wei-etal-2025-plangenllms} remains a significant unresolved bottleneck.

To address these challenges, we propose \Benchmark, which decouples logical pathways from linguistic narratives to enable verifiable evaluation under realistic ambiguity. We further introduce the \Agent-\AndroidA~architecture, which coordinates macro-level context evolution with micro-level hybrid perception to ensure both long-horizon planning consistency and precise on-device execution.

\section{\Benchmark}
\Benchmark~evaluates robustness by decoupling logical structure from linguistic presentation. We employ a three-stage pipeline to instantiate interaction "Naturalness" across five dimensions—Long-horizon Chaining, Cross-domain Grounding, Cognitive Non-linearity, Informational Redundancy, and Contextual Dependency—unifying logical rigor with realistic ambiguity.

\subsection{Methodology for Building Naturalized Causal Pathways}
\begin{figure}
    \centering
    \includegraphics[width=1\linewidth]{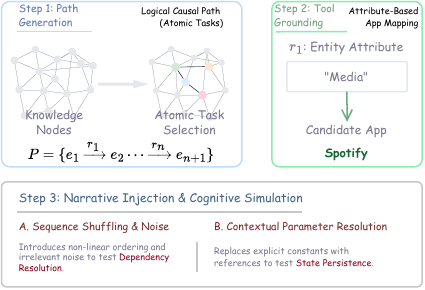}
    \caption{NaturalGAIA build process. This is a three-stage build process, namely Path Generation, Tool Grounding, and Narrative Injection \& Cognitive Simulation.}
    \label{fig:BG}
\end{figure}

As shown in the figure~\ref{fig:BG}. To ensure scalability, we replace ad-hoc manual creation with a systematic process addressing the dimensions above.

\paragraph{Path Generation and Atomic Task Definition (Long-horizon Chaining)}
We formalize a Causal Path (CP) as a traversal sequence $P = \{e_1 \xrightarrow{r_1} e_2 \dots \xrightarrow{r_n} e_{n+1}\}$ on a knowledge graph. Each step, retrieving $e_{i+1}$ from $e_i$ via $r_i$, is an \textit{atomic task}—the smallest interaction unit—where $e_{i+1}$ is the deterministic ground truth. By instantiating dependencies via structured knowledge graphs (e.g., Wikidata), we prevent overfitting and enable \textbf{long-horizon reasoning}, ensuring every task has a verifiable solution grounded in real-world data.

\paragraph{Attribute-Based Tool Mapping (Cross-domain Grounding)}
To address cross-domain scenarios, we map atomic tasks to applications based on target entity attributes. For instance, a task retrieving music metadata is mapped to candidate apps (e.g., Spotify) via the "media" attribute. This separates logical dependencies from interface implementations, ensuring \Benchmark~evaluates core reasoning rather than interface memorization.

\paragraph{Narrative Injection and Cognitive Simulation}
To simulate the remaining cognitive dimensions, we wrap rigid CPs in a natural language layer introducing two key challenges:

\textit{\textbf{Non-Linear Sequencing \& Noise Filtering:}} Human intent rarely adheres to a strict logical sequence. We simulate this by shuffling the sequence of atomic intents and injecting irrelevant noise. This compels the agent to filter noise from the disordered input and reconstruct the valid execution topology, thereby testing its capacity for \textit{dependency resolution}.

\textit{\textbf{Context-Dependent Parameter Resolution:}} We introduce referential ambiguity by replacing explicit constants with context-bound references. This necessitates dynamic context grounding: the agent must recognize that operation targets are anchored to prior execution outputs, requiring a persistent state to resolve abstract pointers into concrete parameters.

\begin{figure*}[t]
    \centering
    \includegraphics[width=0.9\textwidth]{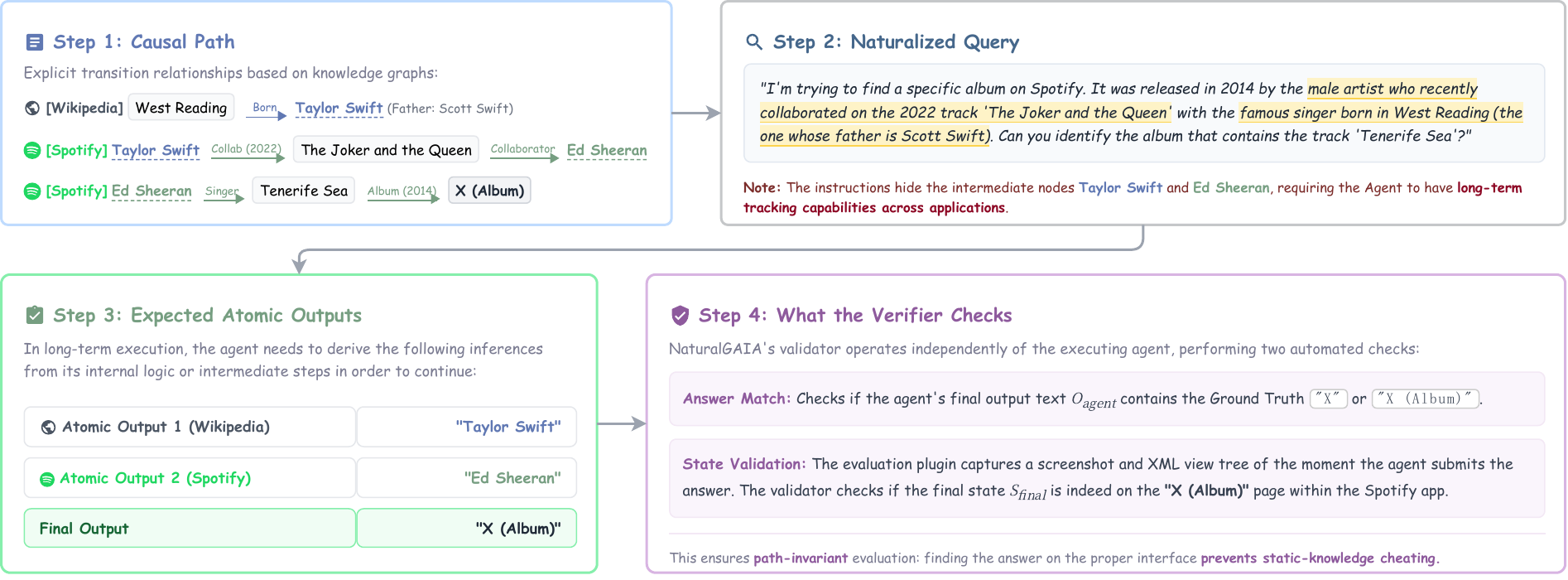}
    \caption{An end-to-end running example illustrating the \Benchmark~pipeline: (i) Causal Path extraction, (ii) Naturalized Query generation simulating cognitive ambiguity, (iii) Expected Atomic Outputs required for progression, and (iv) What the Verifier Checks to ensure path-invariant evaluation.}
    \label{fig:BG2}
\end{figure*}

\textbf{Running Example:} Figure~\ref{fig:BG2} illustrates our end-to-end pipeline. By extracting a deterministic causal path and wrapping it into a Naturalized Query that conceals intermediate nodes, we force the agent to implicitly resolve hidden entities to reach the final verifiable goal.

The specific building principles and processes for \Benchmark~are given in the appendices~\ref{append:Design principles} and~\ref{append:Details of task creation}.

\subsection{Multi-level Evaluation Framework}
To precisely evaluate agent performance within \Benchmark~and meticulously identify the root causes of errors, this study constructs a hierarchical evaluation framework. This framework enables comprehensive scrutiny from macroscopic, difficulty-weighted task completion down to microscopic error attribution.

\paragraph{Level 1: Weighted Pathway Success Rate (WPSR)}
To account for task complexity, we introduce a difficulty score \textbf{$D_{j,i}$} for each task instance $i$ of a Causal Pathway $\text{CP}_j$. This score is proportional to the number of nodes in the pathway and the number of distinct applications involved in its corresponding task. WPSR is defined by weighting each successful completion (\textbf{$\mathbb{S}_{\text{task}}(j,i)=1$}) by its normalized difficulty \textbf{$w_{j,i} = D_{j,i} / \sum_{k,l} D_{k,l}$}, where the sum in the denominator is over all task instances across all pathways.
\begin{equation}
    \text{WPSR} = \sum_{j,i} \textbf{$w_{j,i}$} \cdot \textbf{$\mathbb{S}_{\text{task}}(j,i)$}
\end{equation}

WPSR serves as a holistic metric for an agent's final task completion capability, weighted by difficulty.

\paragraph{Level 2: Fine-grained Traversal Metrics}
To quantify partial progress and resilience, we introduce two complementary metrics that assess traversal quality.

\textbf{Mean Atomic Tasks Completion Ratio (MATCR):} This metric quantifies an agent's ability to successfully complete sequences of atomic tasks. For each task sequence $j$ within the benchmark, a completion ratio $R_j = k_j/n_j$ is calculated, where $k_j$ is the number of consecutive atomic tasks successfully executed from the start, and $n_j$ is the total number of atomic tasks in sequence $j$.
\begin{equation}
    \text{MATCR} = \frac{1}{N} \sum_{j=1}^{N} \frac{k_j}{n_j}
\end{equation}

\textbf{Positional-Weighted Atomic Tasks Success Rate (p-ATSR):} 
This metric evaluates an agent's ability to maintain long-term coherence by assigning greater weight to successes in the later stages of a Causal Pathway. Let $n_j$ be the total number of atomic tasks in pathway $\text{CP}_j$. We introduce a positional weight, denoted as $p(i)$, which is a monotonically increasing function of the step index $i$. The p-ATSR is then defined as:
\begin{equation}
    \text{p-ATSR} = \frac{\sum_{j=1}^{N} \sum_{i=1}^{n_j} p(i) \cdot \mathbb{S}_{\text{atomic}}(j, i)}{\sum_{j=1}^{N} \sum_{i=1}^{n_j} p(i)}
\end{equation}
where $\mathbb{S}_{\text{atomic}}(j, i)=1$ for a success at step $i$ of pathway $j$, and 0 otherwise.

In summary, MATCR assesses an agent's foundational reliability by quantifying its average execution length, whereas p-ATSR places greater emphasis on its long-term coherence by rewarding success in the later stages of a task.

\paragraph{Level 3: Error Attribution Analysis}
For atomic tasks that fail, as identified by the Level 2 metrics, this stage involves analyzing the atomic operation sequence to attribute Execution Errors (EE) to one of three primary types:

\textit{Knowledge Deficit (KD)}: Lack of domain/procedural knowledge for the operation.

\textit{Perceptual Error (PE)}: Correct intent but failed information extraction from the UI.

\textit{Operational Error (OE)}: Correct perception but imprecise action execution.

\subsection{Task Difficulty Stratification}
To enable systematic evaluation, we stratify tasks into three levels based on the topological complexity of their underlying Causal Pathways, specifically considering path length (number of nodes) and cross-application dependencies.

\textit{\textbf{Level 1 (Basic):}} Short pathways (1-2 nodes) confined to 1-2 applications, testing basic sequential execution.

\textit{\textbf{Level 2 (Intermediate):}} Pathways of 3-4 nodes requiring transitions across 3-5 applications, challenging context maintenance.

\textit{\textbf{Level 3 (Advanced):}} Long pathways (5-7 nodes) spanning up to 7 applications. These require advanced planning and memory to manage intricate inter-node dependencies.

Finally, we constructed 276 tasks. Detailed statistics and examples are provided in Appendix~\ref{append:Details of Data Statistics} and~\ref{append:Task example}.

\section{The \Agent-\AndroidA~Collaborative Framework}

To address the complex semantic dependencies and cross-platform barriers inherent in long-horizon tasks, we propose the collaborative \Agent-\AndroidA~architecture. This framework adopts a "macro-planning-micro-execution" paradigm: \Agent~functions as the system's macro-hub responsible for global topology planning, cross-app scheduling, and long-term memory management; while \AndroidA~is designed as a specialized micro-kernel for high-precision instruction grounding on mobile devices.

\subsection{\Agent: Macro-Scheduling and Memory-Driven Flow Management}
\Agent~aims to bridge the gap between abstract user intents and concrete device operations. It serves not only as an initial task decomposer but also as a dynamic scheduling hub, maintaining global context memory and adjusting subsequent plans in real-time based on subtask feedback.

\paragraph{Semantic Parsing and Topology Generation}
The core of the perception layer is the Task Parser ($\mathcal{P}_{\theta}$). Faced with long-horizon user intents $I_{user}$—which often imply complex logical structures such as strict sequential dependencies, parallel execution, or cross-application operations—we model $\mathcal{P}_{\theta}$ as a constrained planning generator. Based on the capability library $\mathbb{A}$, it deconstructs unstructured $I_{user}$ into a logically complete task topology $\mathcal{T}$:
\begin{equation}
    \mathcal{T} = \mathcal{P}(I_{user}, \mathbb{A}) = \langle \tau_1, \tau_2, \dots, \tau_N \rangle
\end{equation}
Each atomic task $\tau_i$ is formalized as a quintuple $\langle \mathcal{D}_i, \mathcal{C}_i, \alpha^*_i, \xi_i, \sigma_i \rangle$, providing the downstream execution with the operation descriptor, context dependency slot, optimal executor route, runtime environment, and lifecycle state, respectively.

\paragraph{Memory-Driven Context Evolution}
At the execution layer, to address information fragmentation caused by heterogeneous environment isolation in long-term tasks, \Agent~abandons static linear execution in favor of an "Execute-Perceive-Evolve" closed-loop strategy.

\textit{\textbf{Context Evolution Mechanism}}: Traditional static instructions fail when subsequent steps depend heavily on the output of preceding ones (e.g., using prior search results as current query keywords). Therefore, we introduce a Context Evolution module $\mathcal{E}_{\phi}$ that is tightly coupled with the system's \textit{global memory}. Before scheduling task $\tau_{i+1}$, the system injects the execution feedback $r_i$ from the preceding task $\tau_i$ into the global memory and dynamically updates the original description of $\tau_{i+1}$, generating a semantically enhanced descriptor $\mathcal{D}'_{i+1}$:
\begin{equation}
    \mathcal{D}'_{i+1} \leftarrow \mathcal{E}_{\phi}(\tau_{i+1}.\mathcal{D}_{raw}, r_i)
\end{equation}
This mechanism enables implicit parameter passing, transforming unstructured natural language feedback from upstream into executable structured parameters downstream, effectively suppressing semantic drift during long-sequence execution. The final system response $\Omega$ is a recursive aggregation process driven by real-time feedback, supplemented by a consistency check $\mathcal{V}(r_t)$ to ensure evidence reliability.

\subsection{\AndroidA: High-Efficiency Android GUI Execution Agent}
As the execution core on mobile devices, \AndroidA~is designed to address the common problems of \textbf{high inference latency} and severe \textbf{coordinate illusions} in existing pure-vision solutions. Through hybrid perception and structured mapping, \AndroidA~achieves efficient and robust single-agent execution.


\paragraph{Hybrid Visual-Structural Perception}
To overcome the limitations of pure vision while preserving visual semantics, the Observer module employs a hybrid strategy to construct the observation $o_t$:
\begin{equation}
    o_t = \langle \mathcal{V}(I_t), \mathcal{T}(X_t) \rangle
\end{equation}
Specifically, $\mathcal{V}(I_t)$ processes the adaptively compressed screenshot $I_t$ to capture unstructured visual semantics with low token overhead. Meanwhile, to eliminate coordinate illusions, $\mathcal{T}(X_t)$ leverages the \textbf{Android Accessibility interface} to retrieve the underlying layout tree $X_t$, pruning it via a viewport semantic filter $S_t = \text{Nodes}(X_t) \cap \mathcal{V}_{\text{vis}} \cap \mathcal{I}_{\text{int}}$. The retained elements are mapped to quadruples (UID, Class, Text, Bounds), where the unique UID provides an unambiguous anchor for the LLM, significantly improving operation accuracy.

\paragraph{Reasoning and Atomic Action Generation}
The Agent Core is driven by a multimodal Large Language Model (LLM). The input prompt $\mathcal{P}_t = \mathcal{I} \oplus h_{t-1} \oplus o_t$ integrates the context-enhanced instruction $\mathcal{I}$, historical trajectory memory $h_{t-1}$, and the current hybrid observation.

\textit{\textbf{Chain-of-Thought Decision Making}}: \AndroidA~enforces an explicit Chain-of-Thought (CoT) mechanism. The policy function $\pi_\theta$ first generates a natural language reasoning path $th_t$, analyzing the gap between the current interface state and the task objective, and then generates standardized action instructions $a_t$:
\begin{equation}
    (th_t, a_t) \sim \pi_\theta(\cdot \mid \mathcal{P}_t)
\end{equation}

\textit{\textbf{Atomic Action Space}}: To ensure execution determinism, we map semantic actions $a_t$ to underlying ADB control signals. The defined atomic action space $\mathcal{A}$ includes:
\begin{itemize}
    \item $\mathtt{TAP}(u)$: Precise clicking based on UID coordinate resolution, completely avoiding pixel prediction errors;
    \item $\mathtt{INPUT}(u, \text{txt})$: A hybrid input strategy that prioritizes simulating a physical keyboard to trigger predictive text, while reverting to broadcast injection for special characters;
    \item $\mathtt{SWIPE/DRAG}$ for gesture control and $\mathtt{SYS}(k)$ for system-level navigation.
\end{itemize}
In summary, Jarvis achieves a precise mapping from abstract instructions to physical signals through a closed loop of $\text{Perception} \rightarrow \text{Reasoning} \rightarrow \text{Action}$:
\begin{equation}
\begin{aligned}
    (\hat{th}, \hat{a}) &= \underset{(th, a)}{\text{argmax}} \, P_{\theta}\left( (th, a) \mid \mathcal{P}_t \right), \\
    a_t^{\text{phy}} &= \mathcal{E}_{\text{act}}\big( (\hat{th}, \hat{a}), \phi(X_t) \big)
\end{aligned}
\end{equation}

We present a complete execution trajectory in the Appendix~\ref{append:ET-Example}.

\section{Experiment}
\begin{table}[h]
    \centering
    \small
    \begin{tabular}{llc}
        \toprule
        \textbf{Baseline} & \textbf{Env} & \textbf{Obs Space} \\
        \midrule
        PCA & Desktop & Scrn (Vision) \\
        MAe & Android & Scrn (Vision) \\
        UI-TARS & Android & Scrn + XML (Hybrid) \\
        \Agent-MAe & Android & Scrn (Vision) \\
        \Agent-\AndroidA & Android & Scrn + XML (Hybrid) \\
        \bottomrule
    \end{tabular}
    \caption{Comparison of observation spaces across baselines.}
    \label{tab:observation_space}
\end{table}

\begin{table*}[!ht]
\centering
\setlength{\tabcolsep}{1.5pt}
\renewcommand{\arraystretch}{1.15}
\scriptsize
\scalebox{1.1}{
\begin{tabular}{l cccc cccc cccc cccc}
\toprule[1.5pt]
\multirow{3}{*}{Method} 
& \multicolumn{4}{c}{Level-1} 
& \multicolumn{4}{c}{Level-2} 
& \multicolumn{4}{c}{Level-3} 
& \multicolumn{4}{c}{Overall} \\
\cmidrule(lr){2-5} \cmidrule(lr){6-9} \cmidrule(lr){10-13} \cmidrule(lr){14-17}

& SR & \multirow{2}{*}{\shortstack{WPSR}} & \multirow{2}{*}{\shortstack{MAT\\CR}} & \multirow{2}{*}{\shortstack{ATSR}} 
& SR & \multirow{2}{*}{\shortstack{WPSR}} & \multirow{2}{*}{\shortstack{MAT\\CR}} & \multirow{2}{*}{\shortstack{ATSR}}
& SR & \multirow{2}{*}{\shortstack{WPSR}} & \multirow{2}{*}{\shortstack{MAT\\CR}} & \multirow{2}{*}{\shortstack{ATSR}}
& SR & \multirow{2}{*}{\shortstack{WPSR}} & \multirow{2}{*}{\shortstack{MAT\\CR}} & \multirow{2}{*}{\shortstack{ATSR}} \\

& (P@1/4) & & & 
& (P@1/4) & & & 
& (P@1/4) & & & 
& (P@1/4) & & & \\
\midrule
\multicolumn{17}{l}{\cellcolor{gray!15}\textbf{PC-Agent}} \\ 
Gemini-2.5-Pro & 40.0 / 66.7 & 42.9 & 60.8 & 59.1 & 10.0 / 40.0 & 10.0 & 43.1 & 32.0 & 0.0 / 20.0 & 7.6 & 24.9 & 15.9 & 20.0 / 45.7 & 13.1 & 45.5 & 25.7 \\ \midrule
\multicolumn{17}{l}{\cellcolor{gray!15}\textbf{Mobile-Agent-e}} \\ 
Gemini-2.5-Pro & 46.7 / \textbf{100.0} & 58.9 & 74.2 & 68.9 & 10.0 / \underline{60.0} & 20.0 & 50.0 & 41.0 & 0.0 / 30.0 & 12.7 & 24.4 & 17.5 & 22.9 / 68.6 & 21.1 & 53.0 & 30.4 \\
Gemini-2.5-flash & 40.0 / 80.0 & 46.4 & 60.0 & 56.1 & 0.0 / 50.0 & 15.0 & 42.5 & 33.0 & 0.0 / 20.0 & 12.7 & 15.0 & 13.8 & 17.1 / 54.3 & 18.0 & 42.1 & 24.4 \\ \midrule
\multicolumn{17}{l}{\cellcolor{gray!15}\textbf{LightManus\_Mobile-Agent-e}} \\ 
Gemini-2.5-Pro & \underline{73.3} / \textbf{100.0} & 63.4 & 71.7 & 71.3 & 20.0 / \textbf{70.0} & 27.5 & 53.1 & 44.5 & 10.0 / 40.0 & 20.3 & 29.9 & 25.2 & 40.0 / 74.3 & 28.3 & 54.4 & 36.3 \\
Gemini-2.5-flash & 53.3 / 80.0 & 63.4 & 74.2 & 71.3 & 0.0 / \underline{60.0} & 22.5 & 50.0 & 42.0 & 0.0 / 30.0 & 12.7 & 20.5 & 16.1 & 22.9 / 60.0 & 22.5 & 51.9 & 30.1 \\
GPT-5.2 & 60.0 / - & 57.1 & 66.7 & 61.0 & 30.0 / - & 30.0 & 30.0 & 30.0 & 20.0 / - & 18.6 & 20.0 & 17.6 & 40.0 / - & 27.2 & 42.9 & 26.4 \\
Claude-Sonnet-4.5 & 60.0 / - & 57.1 & 60.0 & 56.1 & \underline{40.0} / - & 40.0 & 57.5 & 49.0 & 20.0 / - & 20.3 & 36.7 & 28.4 & 42.9 / - & 31.1 & 52.6 & 37.7 \\ \midrule
\multicolumn{17}{l}{\cellcolor{gray!15}\textbf{LightManus\_Jarvis}} \\ 
Gemini-2.5-pro & \underline{73.3} / \textbf{100.0} & \underline{75.0} & 78.3 & 77.4 & \underline{40.0} / \textbf{70.0} & 42.5 & \underline{70.6} & 63.0 & 20.0 / \underline{50.0} & 27.5 & 50.8 & 42.2 & 48.6 / \underline{77.1} & 38.3 & \underline{68.3} & 52.4 \\
Gemini-2.5-flash & 66.7 / \underline{93.3} & 69.6 & 75.8 & 73.8 & 30.0 / \underline{60.0} & 37.5 & 65.6 & 57.0 & 10.0 / 30.0 & 20.3 & 32.5 & 27.5 & 40.0 / 65.7 & 32.0 & 60.5 & 41.5 \\
Gemini-3.0-pro & \underline{73.3} / \textbf{100.0} & \textbf{81.2} & \underline{83.3} & \underline{82.3} & \underline{40.0} / \textbf{70.0} & 45.0 & \textbf{75.0} & \textbf{67.5} & \underline{30.0} / \textbf{60.0} & 34.7 & \underline{55.5} & \textbf{46.7} & \underline{51.4} / \textbf{80.0} & \underline{44.1} & \textbf{73.0} & \textbf{57.0} \\
Gemini-3.0-flash & \textbf{86.7} / \underline{93.3} & \textbf{81.2} & \textbf{86.7} & \textbf{84.8} & 30.0 / \textbf{70.0} & \underline{47.5} & 68.8 & 60.8 & \underline{30.0} / 40.0 & 27.1 & 40.1 & 32.2 & \textbf{54.3} / 71.4 & 40.4 & 68.2 & 46.7 \\
GPT-5.2 & 66.7 / - & 64.3 & 76.7 & 70.7 & \underline{40.0} / - & 40.0 & 40.0 & 40.0 & \textbf{40.0} / - & \textbf{40.7} & 40.0 & 41.2 & \underline{51.4} / - & 43.7 & 55.7 & 44.3 \\
Claude-Sonnet-4.5 & 66.7 / - & 64.3 & 73.3 & 68.3 & \textbf{50.0} / - & \textbf{50.0} & 70.0 & \underline{65.0} & \textbf{40.0} / - & \underline{39.0} & \textbf{56.7} & \underline{45.6} & \textbf{54.3} / - & \textbf{45.6} & 67.6 & \underline{53.9} \\
Qwen3-max & 53.3 / 86.7 & 58.0 & 62.5 & 61.0 & 30.0 / \textbf{70.0} & 37.5 & 39.4 & 38.5 & 10.0 / 30.0 & 20.3 & 20.0 & 20.6 & 34.3 / 65.7 & 30.5 & 43.8 & 30.6 \\
Qwen3-vl-plus & 46.7 / 73.3 & 45.5 & 55.8 & 53.7 & 20.0 / 40.0 & 27.5 & 43.8 & 38.0 & 10.0 / 30.0 & 12.7 & 18.9 & 15.2 & 28.6 / 51.4 & 21.5 & 41.8 & 26.4 \\
Qwen3.5-397b-a17b & 60.0 / \underline{93.3} & 59.8 & 64.2 & 62.8 & 30.0 / \underline{60.0} & 37.5 & 46.2 & 43.8 & 20.0 / 40.0 & 22.9 & 23.8 & 23.9 & 40.0 / 74.3 & 32.2 & 47.5 & 34.3 \\
UI-TARS-1.5-7b & 33.3 / 80.0 & 42.0 & 56.7 & 53.0 & 0.0 / 30.0 & 10.0 & 41.2 & 31.0 & 0.0 / 20.0 & 7.6 & 14.2 & 12.4 & 14.3 / 48.6 & 13.0 & 40.1 & 22.6 \\
\bottomrule[1.5pt]
\end{tabular}
}
\caption{Main results on the \Benchmark. SR (P@1/4) denotes Success Rate at Pass@1 and Pass@4. \textbf{Bold} indicates the best performance, and \underline{underline} indicates the second best. Note that for Claude and GPT series, only Pass@1 is reported due to computational constraints.}
\label{tab:main_result_optimized}
\end{table*}


\begin{table}[t]
\centering
\setlength{\tabcolsep}{4pt}
\renewcommand{\arraystretch}{1.15}
\scriptsize
\scalebox{1.0}{
\begin{tabular}{l cccc}
\toprule[1.5pt]
\multirow{2}{*}{Model} & \multicolumn{3}{c}{Token Usage} & \multirow{2}{*}{Duration (s)} \\
\cmidrule(lr){2-4}
 & Input & Output & Total & \\
\midrule
MAe & 377,200 & 50,780 & 427,980 & 2015.4 \\
\Agent-MAe & 378,300 & 51,850 & 430,150 & 2046.9 \\ 
\Agent-\AndroidA & 98,700 & 13,900 & 112,600 & 505.8 \\
\bottomrule[1.5pt]
\end{tabular}
}
\caption{Efficiency analysis on long-horizon Level-3 tasks using Gemini-3.0-Flash. The reported values represent the average metrics across 5 identical L3 tasks (averaging 5.4 atomic tasks).}
\label{tab:agent_performance_formatted}
\end{table}

\subsection{Setup}
\paragraph{Experimental Environment.} Experiments are executed on an Ubuntu 20.04 server. The Android testing environment is an Android Studio emulator (Android 11.0, 6 vCPUs, 8 GB RAM). The Windows environment is a VMware virtual machine (Windows 11, 8 vCPUs, 12 GB RAM).



\paragraph{Baselines \& Observation Spaces.} We compare \Agent~with PC-Agent (PCA)~\cite{he2024pcagent}, Mobile-Agent-e~(MAe)~\cite{wang2025mobile}, and UI-TARS~\cite{qin2025uitarspioneeringautomatedgui}. An ablation variant, \Agent-MAe, isolates the macro-planner's contribution. As Table~\ref{tab:observation_space}, our independent verifier strictly enforces a unified State Validation standard (capturing screenshots and XML) across all models to ensure an absolutely consistent evaluation.

\paragraph{Foundation Models.} We test multiple MLLM backends for generalizability: closed-source models (Gemini 2.5/3 Pro/Flash, GPT-5.2, Claude-Sonnet-4.5) and open-source models (Qwen3-max, Qwen3-plus, and Qwen3.5-397b-a17b).

\subsection{Main Results}

Table~\ref{tab:main_result_optimized} presents the experimental results of different agent architectures and foundation models on \Benchmark. Overall, the proposed \Agent-\AndroidA~architecture significantly outperforms baseline methods across all metrics. The results not only validate the effectiveness of the architecture itself but also reveal complex interactions between model capabilities and task difficulty.

\paragraph{Architecture Effectiveness: Asymmetry of Gains}
Our empirical results strongly corroborate the superiority of the "macro-planning-micro-execution" paradigm. Using Gemini-2.5-Pro as the foundation model, merely introducing the \Agent~scheduler boosts the WPSR of MAe from 21.1\% to 28.3\%; further integrating the \AndroidA~executor yields a significant leap to 38.3\%. This asymmetry indicates that optimizing planning alone fails to resolve perceptual hallucinations; \AndroidA's high-precision atomic operations are crucial for success.

Regarding long-horizon coherence (measured by p-ATSR), \Agent-\AndroidA~(52.4\%) also significantly outperforms MAe (30.4\%). A more compelling finding is that even with the lightweight Gemini-2.5-Flash, our architecture's p-ATSR (41.5\%) surpasses MAe equipped with a stronger foundation model. This proves that our architecture maintains long-sequence context consistency, effectively compensating for the limitations of weaker foundation models.

\paragraph{Model Performance: Task-Driven Divergence}
The performance of different models does not scale linearly with parameter size but exhibits a divergence trend highly correlated with task characteristics.

Lightweight models (e.g., Gemini3-Flash) demonstrate superior immediate instruction-following capabilities in Level-1 basic tasks (WPSR 81.2\%), even outperforming the larger Gemini-3.0-Pro. Our analysis suggests that for short-horizon, linear tasks, efficient intuitive instruction mapping is more effective than complex logical scrutiny, as the latter can sometimes lead to "over-interpretation" of simple directives.

However, when facing Level-3 complex tasks laden with cognitive non-linearity and cross-domain dependencies, models with strong logical reasoning capabilities (e.g., Claude-Sonnet-4.5 and Gemini-3.0-Pro) exhibit irreplaceable robustness. For instance, while Claude-Sonnet-4.5 performed typically in Level-1 due to hypersensitivity to instruction noise, its powerful logical deduction and noise filtering abilities enabled it to achieve the highest Mean Atomic Tasks Completion Ratio (MATCR 56.7\%) in Level-3. By examining execution logs, we found that while some open-source models (e.g., Qwen3-Max) possess reliable single-step execution skills, their long-horizon logical coherence decays as steps increase, making them struggle with complex contextual dependencies.Similarly, while UI-TARS-1.5-7B performs adequately on Level-1, its Pass@1 drops to 0\% on Level-3, directly exposing its vulnerabilities in long-horizon logic decay.

\paragraph{Task Difficulty and Error Propagation}
Performance declines non-linearly as task difficulty rises, validating the challenges introduced by the "Naturalness" of \Benchmark. For instance, Gemini-3.0-Pro's WPSR drops from 81.2\% (Level-1) to 34.7\% (Level-3). However, its relatively high MATCR (55.5\%) in Level-3 reveals a disparity between local step execution and global task completion. We attribute this failure to two distinct factors consistent with our framework: (1) \textit{Intent Parsing Failures}, where models struggle to decompose ambiguous, natural language instructions into correct planning topologies; and (2) \textit{Cumulative Error Propagation}, where minor deviations in atomic operations accumulate over long horizons, causing the execution chain to fracture. This highlights that high success rates depend on both precise semantic disambiguation and robust error-correction mechanisms.

\paragraph{Supplementary Experiments.} To validate generalizability, we evaluated \Agent-\AndroidA~on the independent AndroidWorld benchmark and compared it against Mind2Web on a pure-web subset (details in Appendix~\ref{append:benchmark_comparison}).

\subsection{Error analysis}
\begin{figure}[!h]
    \centering
    \includegraphics[width=1\linewidth]{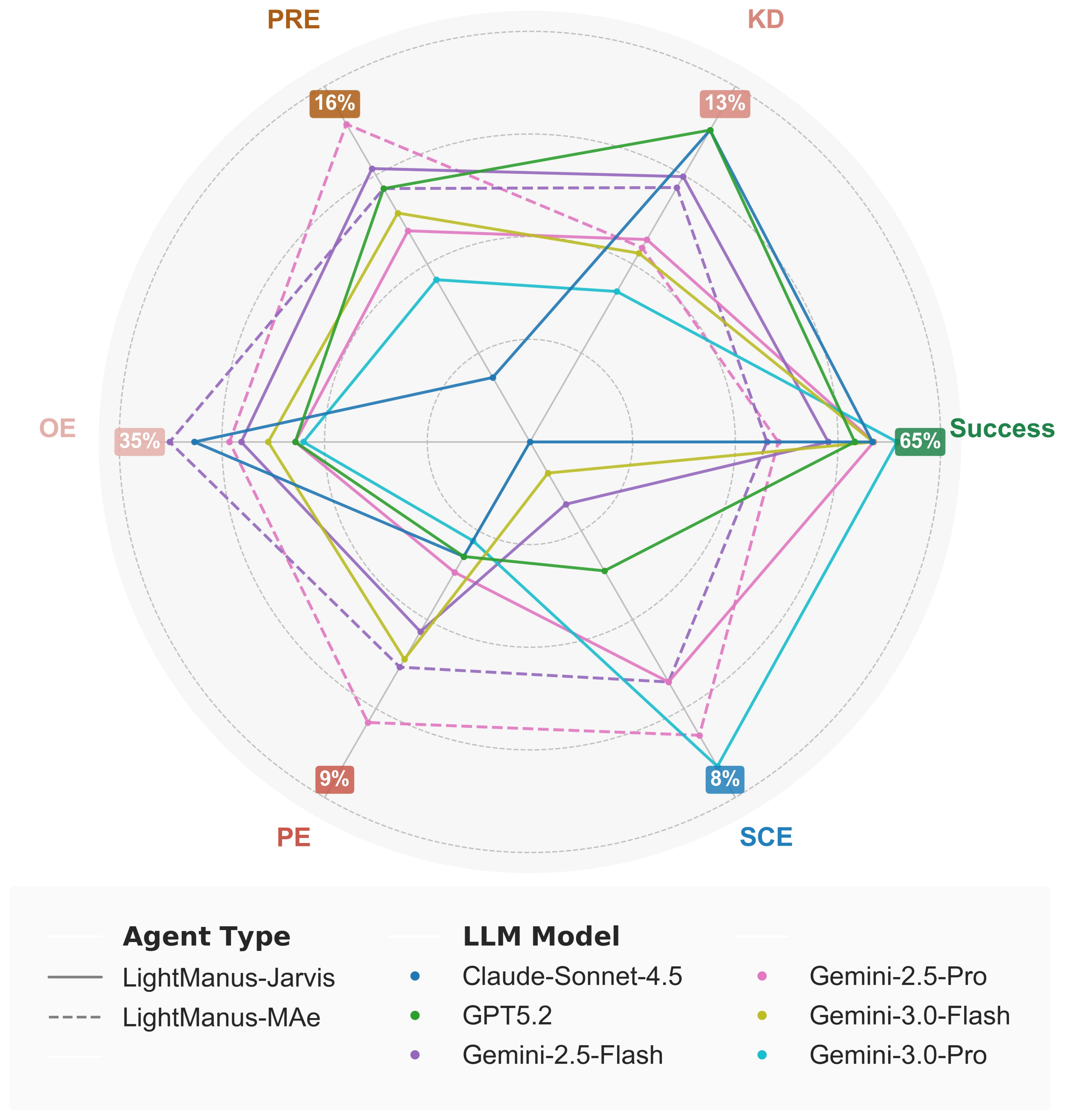}
    \caption{A comparative analysis of the failure reasons of different GUI agents and LLM backends. The chart categorizes errors into three main types: Knowledge Deficit (KD), Planning \& Reasoning Errors (PRE), Operational Errors (OE), Perceptual Errors (PE), and Structural Compliance Errors (SCE)}
    \label{fig:model_error_analysis}
\end{figure}

To analyze the bottlenecks limiting agent performance, we categorize failure cases into five types: Knowledge Deficit (KD), Planning \& Reasoning Errors (PRE), Operational Errors (OE), Perceptual Errors (PE), and Structural Compliance Errors (SCE). Figure~\ref{fig:model_error_analysis} and the detailed statistics reveal three key observations. We have provided specific examples of several errors in the Appendix~\ref{append:Error Type}.

\paragraph{Execution Remains the Primary Bottleneck}
Across all architectures, OE consistently constitutes the primary failure mode, ranging from 19\% to 30\% even in advanced models like Claude-Sonnet-4.5. This bottleneck stems not merely from layout variability, but from the dynamic instability of mobile environments. First, high-density GUI contexts often induce visual hallucinations regarding element coordinates. Second, a temporal mismatch exists between reasoning and execution: the inherent latency of LLM inference allows interface states to drift (e.g., sudden pop-ups) before the action is applied. Furthermore, reliance on discrete static observations rather than real-time monitoring leaves agents vulnerable to transient state changes, resulting in precise actions being executed on obsolete states.

\paragraph{Efficacy of the Specialized Execution Module}
Comparing \Agent-MAe with \Agent-\AndroidA~validates the design of our \AndroidA~micro-kernel. For Gemini-2.5-Pro, upgrading to the Jarvis architecture reduced PE from 7.1\% to 3.3\% and OE from 25.6\% to 20.0\%. This reduction demonstrates that the hybrid visual-structural perception mechanism in Jarvis effectively mitigates "coordinate illusions" and improves the accuracy of element localization, confirming that a specialized execution agent is superior to general-purpose visual execution.

\paragraph{The Reasoning-Execution Trade-off}
We observe a distinct divergence between abstract planning and concrete execution. Claude-Sonnet-4.5, despite achieving minimal PRE (2.9\%), struggles with high OE and KD (11.4\%). In contrast, the Gemini-3.0 series maintains a more robust balance. Crucially, we find that KD is inversely correlated with model scale. Larger models leverage extensive parametric memory to recall diverse app-specific workflows, evidenced by the lower KD in Gemini-3.0-Pro (5.5\%) compared to its Flash counterpart (6.9\%). This suggests that robust agent performance relies on the synergy of two distinct capabilities: logical reasoning for topological planning and scale-driven domain knowledge for procedural grounding.


\subsection{Efficiency Analysis}
Beyond accuracy metrics, we evaluate the operational efficiency of the proposed framework. As presented in Table~\ref{tab:agent_performance_formatted}, when handling the most complex long-sequence Level-3 tasks, \AndroidA~demonstrates superior efficiency compared to the GUI agent MAe. Specifically, Jarvis reduces total token consumption by approximately 74\% (112,600 vs. 427,980) and shortens task duration by 75\% (505.8s vs. 2015.4s). Furthermore, compared to MAe, the ablation variant \Agent-MAe introduces negligible overhead (+31.5s, +1.1k tokens), proving that our macro-planner provides global context management with extremely low computational cost. By replacing redundant visual encoding with a lightweight hybrid perception mechanism, our approach significantly minimizes inference latency and deployment costs.

\section{Conclusion and Discussion}
This study highlights the "Evaluation-Realism Dilemma" constraining GUI agent research. By introducing \Benchmark, we provide a methodology to decouple logical rigidity from linguistic naturalness. Furthermore, the \Agent-\AndroidA~framework validates the efficacy of the "macro-planning-micro-execution" paradigm in mitigating semantic drift and execution noise. Experiments demonstrate that this hierarchical approach significantly enhances robustness and efficiency compared to existing baselines. We hope this work offers valuable insights into the reasoning-execution gap and serves as a foundation for developing more robust, ecologically valid autonomous systems.

\section*{Limitations}
Our work entails limitations in both benchmark construction and agent efficacy. First, the pursuit of verifiable determinism constrains task diversity. To ensure rigorous ground truth, we exclude dynamic or irreversible scenarios (e.g., e-commerce, real-time messaging), which consequently narrows the scope of supported applications compared to open-ended environments. Second, scalability is bottlenecked by manual verification. Although task generation is automated, validating the topological logic currently relies on human experts, as existing agents lack the capability for reliable automated quality assurance. Finally, regarding the agent, inference latency and cumulative error propagation remain unresolved. The temporal mismatch between MLLM inference and dynamic interface changes can lead to state drift, and minor execution deviations in long-horizon tasks still pose risks of cascading failure.

\section*{Acknowledgments}
This work was supported by the Brain Science and Brain-like Intelligence Technology-National Science and Technology Major Project under Grant 2022ZD0208900, the National Natural Science Foundation of China under Grant 52308250, and the Guangdong Basic and Applied Basic Research Foundation under grant 2026A1515012965.


\bibliography{custom}

\appendix
\section{Design Principles for \Benchmark}
\label{append:Design principles}

To ensure the benchmark rigorously evaluates agentic capabilities in realistic scenarios, we formulate a set of strict design principles governing the generation of high-complexity tasks (referred to as Level-3 complexity). These principles enforce structural integrity, logical determinism, and ecological validity.

\paragraph{I. Structural Complexity and Cross-Domain Scope}
\begin{itemize}
    \item \textbf{Long-Horizon Composition:} To challenge the agent's context maintenance capabilities, each main task must be composed of a sequence of atomic interactions (typically $N \ge 6$). These atomic tasks are explicitly delimited in the logical skeleton, preventing ambiguous task boundaries.
    \item \textbf{Multi-App Collaboration:} The task execution must necessitate context switching between distinct application environments. We enforce a minimum diversity constraint (e.g., interacting with at least three distinct applications), ensuring the agent is evaluated on cross-domain grounding rather than single-app proficiency.
\end{itemize}

\paragraph{II. Determinism of Atomic Outcomes}
\begin{itemize}
    \item \textbf{Explicit Textual Output:} Every atomic task must yield a verifiable result presentable in text format. This enables automated evaluation of intermediate steps.
    \item \textbf{Uniqueness and Objectivity:} The output space of each atomic task must be constrained to a single, unambiguous ground truth (Deterministic Causality). Ambiguous queries yielding multiple valid answers are strictly pruned during the generation phase.
    \item \textbf{Temporal Stability and Freeze-Frame Strategy:} To ensure reproducibility, tasks prioritize time-invariant knowledge (e.g., historical data, physical constants). For scenarios inevitably involving time-sensitive information (e.g., "current stock price"), we employ a \textbf{Freeze-Frame Strategy}: the information is solidified at the moment of capture into a local static snapshot (e.g., a cached page or screenshot). The ground truth is then validated against this static context rather than the live web, eliminating drift caused by temporal latency.
\end{itemize}

\paragraph{III. Logical Dependency Chain}
\begin{itemize}
    \item \textbf{Strict State Dependency:} The task structure enforces a rigid sequential dependency, where the output $O_{t}$ of the atomic task at step $t$ serves as a mandatory input parameter for step $t+1$. This creates an unbreakable "information flow."
    \item \textbf{Chain Irreducibility:} The dependency chain is designed such that a failure in any preceding node renders subsequent steps theoretically executing impossible. This "Fail-Stop" mechanism ensures that agents cannot bypass reasoning steps through hallucination or guessing.
\end{itemize}

\paragraph{IV. Grounding Specification}
\begin{itemize}
    \item \textbf{App-Specific Designation:} Each atomic task is explicitly mapped to a specific target application within the defined schema. The benchmark rejects ambiguous instructions that do not imply a clear tool selection.
    \item \textbf{Verifiable Execution Path:} Alongside the final answer, the benchmark records the canonical operation path (the sequence of edges traversed in the knowledge graph) as the Golden Path. This facilitates granular error analysis beyond simple success/failure metrics.
\end{itemize}

\paragraph{V. Ecological Validity of Tools}
\begin{itemize}
    \item \textbf{Popularity-Based Selection:} To ensure the benchmark reflects real-world utility, the selection of applications follows the principle of ecological validity. We prioritize applications with high market penetration (e.g., top-ranked apps in global stores\footnote{Data reference: Google Play Store rankings via \url{https://play.google.com/store}.}), ensuring the agent learns to interact with interfaces relevant to the general user population.
\end{itemize}

\section{Details of Task Creation Pipeline}
\label{append:Details of task creation}

The construction of \Benchmark~is grounded in the semantic structure of Wikidata, utilizing its RDF (Resource Description Framework) triples to instantiate a realistic GUI environment. We formalize the pipeline into six stages: Semantic Substrate Analysis, Information Space Definition, Subgraph Construction, Causal Skeleton Generation, Natural Language Mapping, Human Verification Protocol, and a Case Study.

\subsection{Data Structure and Semantic Substrate}
We select Wikidata as the semantic substrate for our benchmark. Structurally, Wikidata is a labeled multidigraph where the fundamental data unit is an RDF triple $(s, p, o)$, denoting a subject $s$ linked to an object $o$ via a predicate $p$. This structure provides two methodological foundations for task generation:
\begin{enumerate}
    \item \textbf{Deterministic Causality:} Unlike unstructured text, RDF triples provide unambiguous logical transitions. For instance, if a task requires identifying a "director," the predicate $P57$ provides a deterministic edge from a movie entity to a person entity, eliminating ambiguity in logical reasoning.
    \item \textbf{Ontological Isomorphism:} The class hierarchy in Wikidata (defined via $P31$ \textit{instance of} and $P279$ \textit{subclass of}) mirrors real-world taxonomies. This allows us to project the global graph into specific domain contexts using strict ontological rules rather than arbitrary node connections.
\end{enumerate}

\subsection{Definition of Application Information Spaces}
To simulate a multi-application environment, we define an \textbf{Information Space} ($\mathcal{I}_{app}$) for each application. An information space is not merely a collection of entities but a constrained subgraph defined by a tuple $(\mathcal{T}, \mathcal{R}, \mathcal{A})$:
\begin{itemize}
    \item \textbf{Root Classes ($\mathcal{T}$):} The set of entity types allowed within the application (corresponding to \texttt{type\_filter} in configuration). For example, a music app includes classes such as "Musical Recording" and "Artist".
    \item \textbf{Navigable Relations ($\mathcal{R}$):} The specific Wikidata predicates that function as interactive transitions within the GUI (corresponding to \texttt{actions}). An edge in the graph is instantiated as a clickable link only if its predicate belongs to $\mathcal{R}$.
    \item \textbf{Attribute Space ($\mathcal{A}$):} The set of properties visible on an entity's profile page (corresponding to \texttt{constraints}). These attributes serve as metadata for filtering and verification.
\end{itemize}

Figure \ref{fig:config_json} illustrates a configuration snippet for the "Spotify" application. The action \texttt{view\_artist} maps the predicate $P175$ to a transition from a \textit{Track} to an \textit{Artist}, while properties like $P577$ (publication date) constitute the observable attribute space.

\begin{figure}[h]
\centering
\begin{lstlisting}[language=json, basicstyle=\ttfamily\footnotesize, frame=single]
"Spotify": {
  "description": "Music streaming service...",
  "entities": {
    "Track": {
      "type_filter": ["Q218818", "Q7366"], 
      // Root Classes: Musical Recording, Song
      "constraints": ["P175", "P577", "P136"], 
      // Attribute Space: Performer, Date, Genre
      "actions": {
        "view_artist": {
          "target": "Artist", 
          "relation": "P175", 
          // Navigable Relation: P175 (Performer)
          "intent": "find the performing artist"
        }
      }
    }
  }
}
\end{lstlisting}
\caption{Configuration snippet defining the Information Space for the Spotify application.}
\label{fig:config_json}
\end{figure}

A critical property of our design is that the information space of any vertical application is a proper subset of the general knowledge space (e.g., Wikipedia), denoted as $\mathcal{I}_{vertical} \subset \mathcal{I}_{wiki}$. This ensures that Wikipedia acts as a connectivity guarantor: when no direct path exists between atomic tasks in vertical apps, the agent can utilize Wikipedia as a hub to bridge disconnected components, enabling long-horizon task continuity.

\subsection{Subgraph Construction with Type Inference}
We employ a seed-based expansion strategy to construct the evaluation subgraph. To ensure adherence to the defined schema, we implement a deep type inference mechanism during traversal.
Instead of retrieving all neighbors, the crawler verifies the class hierarchy of each neighbor using the SPARQL property path \texttt{wdt:P31/wdt:P279*}. A node is included in the subgraph only if its ancestral classes intersect with the target application's Root Classes ($\mathcal{T}$). This filters out ontological noise and ensures logical consistency within the generated environment.

\subsection{Causal Skeleton Generation and Cardinality Constraints}
We generate the logical skeleton of a task via constrained random walks on the constructed subgraph. A step in the causal path is defined as $S_t = (v_p, e, v'_{p}, \mathbb{C})$, generated as follows:
\begin{enumerate}
    \item \textbf{Primary Node Transition ($v_p \to v'_{p}$):} The primary node represents the focal entity of the current interface (e.g., an Album). The system selects an edge $e$ based on the application's navigable relations $\mathcal{R}$ to transition to the next primary node $v'_{p}$ (e.g., the Artist).
    \item \textbf{Application Switching:} The predicate of edge $e$ determines the next active application. For instance, selecting $P175$ activates the music application, whereas $P57$ activates the movie database.
    \item \textbf{Cardinality Constraints and Secondary Nodes ($\mathbb{C}$):} Relations in Wikidata are often one-to-many (e.g., an artist has multiple albums). To uniquely identify the target node $v'_{p}$, the system retrieves distinctive attributes from the attribute space $\mathcal{A}$ of $v'_{p}$ (e.g., release year, genre). These attributes form the set of Secondary Nodes $\mathbb{C}$.
\end{enumerate}
In the final task, these Secondary Nodes serve as discriminative constraints (e.g., "the pop album released in 2014"). The agent must utilize these visible attributes to filter candidates and resolve branching ambiguities.

\subsection{Natural Language Mapping and Cognitive Simulation}
The logical skeleton is transformed into a natural language instruction using a Large Language Model (LLM). This process follows a "Hide-and-Describe" principle to simulate cognitive challenges:
\begin{itemize}
    \item \textbf{Hiding Intermediate Nodes:} Intermediate primary nodes in the path are redacted from the instruction. The agent is provided only with the start node and the final intent, requiring it to infer the traversal path.
    \item \textbf{Constraint Description:} The discriminative constraints ($\mathbb{C}$) selected in the previous stage are converted into natural language descriptors. For example, the structured constraint $(P577: 2014)$ is rewritten as the relative clause "released in 2014".
    \item \textbf{Narrative Injection:} To simulate diverse user intents, the logical conditions are wrapped in varying narrative templates (e.g., "Exploration," "Vague Recollection"), increasing the pragmatic complexity of the instruction.
\end{itemize}

\subsection{Human Verification Protocol}
\label{append:Human Verification}

While the pipeline described above is automated, ensuring the fidelity of the benchmark requires a Human-in-the-Loop (HITL) verification stage. We employ a rigorous two-step validation protocol for the generated tasks:

\paragraph{Step 1: Solvability Check}
Human annotators act as the agents. They are provided with the initial instruction and the environment constraints (the specific versions of the Apps). Annotators must attempt to solve the task following the generated logical path. Any task where the ground truth cannot be deterministically reached—due to interface updates, data obsolescence, or graph ambiguity—is flagged and discarded.

\paragraph{Step 2: Consistency Verification}
Reviewers inspect the alignment between the generated \textit{Natural Language Query} (from Section A.5) and the \textit{Logical Skeleton} (from Section A.4). They verify that:
\begin{itemize}
    \item The natural language query does not accidentally leak the hidden intermediate entities (Spoiler Check).
    \item The query contains sufficient linguistic cues (e.g., accurate descriptions of constraints) to logically deduce the next step without ambiguity.
\end{itemize}
Only tasks that pass both solvability and consistency checks are included in the final \Benchmark~release.

\subsection{Case Study: From Graph to Natural Instruction}
We illustrate the pipeline with a generated example.

\paragraph{1. Logical Skeleton Generation}
The generator performs a random walk involving Wikipedia and Spotify:
\begin{itemize}
    \item \textbf{Step 1 (Wikipedia): Place $\to$ Person.} 
    Transition: \textit{West Reading} $\xrightarrow{P19^{-1}}$ \textit{Taylor Swift}.
    Constraint ($\mathbb{C}_1$): Father is \textit{Scott Swift}; Occupation is \textit{Singer-songwriter}.
    
    \item \textbf{Step 2 (Spotify): Person $\to$ Track.} 
    Transition: \textit{Taylor Swift} $\xrightarrow{P175^{-1}}$ \textit{The Joker and the Queen}.
    Constraint ($\mathbb{C}_2$): Released in \textit{2022}; Co-artist is \textit{Ed Sheeran}.
    
    \item \textbf{Step 3 (Spotify): Track $\to$ Co-Artist.} 
    Transition: \textit{The Joker and the Queen} $\xrightarrow{P175}$ \textit{Ed Sheeran}.
    Constraint ($\mathbb{C}_3$): Genre includes \textit{Pop music}.
    
    \item \textbf{Step 4 \& 5 (Spotify): Artist $\to$ Track $\to$ Album (Goal).} 
    Transition: \textit{Ed Sheeran} $\to$ \textit{Tenerife Sea} $\to$ \textit{X (Album)}.
    Constraint ($\mathbb{C}_5$): Released in \textit{2014}; Label is \textit{Asylum Records}.
\end{itemize}

\paragraph{2. Naturalization}
The LLM converts the skeleton into a user query, hiding intermediate entities like "Taylor Swift" and "Ed Sheeran":

\begin{quote}
"I'm trying to find a specific album on Spotify. It was released in \textbf{2014} by the male artist who recently collaborated on the 2022 track 'The Joker and the Queen' with the famous singer born in \textbf{West Reading} (the one whose father is \textbf{Scott Swift}). Can you identify the album that contains the track '\textbf{Tenerife Sea}'?"
\end{quote}

\paragraph{3. Resolution Logic}
The agent must resolve the dependency chain: (1) Identify \textit{Taylor Swift} via the "West Reading/Scott Swift" constraint on Wikipedia; (2) Locate the 2022 collaboration to identify \textit{Ed Sheeran}; (3) Navigate to Sheeran's profile to find the 2014 album containing "Tenerife Sea". This demonstrates the benchmark's ability to create rigorous long-horizon tasks requiring cross-app information retrieval and state tracking.

\section{Details of Data Statistics and Environment}
\label{append:Details of Data Statistics}

\subsection{Task Complexity Distribution}
Table \ref{tab: Statistics} presents the statistical characteristics of the \Benchmark~dataset across varying difficulty levels. The dataset is stratified into three levels based on the length of the causal chain. For each stratum, we quantify the average count (Avg) and distribution range of both \textit{Tools Involved} and \textit{Atomic Tasks}. 
The data exhibits a strong positive correlation between difficulty level and structural complexity: Level-3 tasks require, on average, interaction with 4.4 distinct applications and a sequence of 5.7 atomic operations. This distribution confirms that \Benchmark~successfully captures a wide spectrum of complexity, from simple single-step retrievals to long-horizon, cross-domain reasoning chains.

\begin{table*}[h]
\centering
\caption{Statistical Distribution of Task Complexity in \Benchmark}
\label{tab: Statistics}
    \begin{tabular}{l c c c c c}
        \toprule[1.5pt]
        \multirow{2}{*}{\textbf{Difficulty Level}} & \multirow{2}{*}{\textbf{Count}} & \multicolumn{2}{c}{\textbf{Unique Tools}} & \multicolumn{2}{c}{\textbf{Atomic Tasks}} \\
        \cmidrule(lr){3-4} \cmidrule(lr){5-6}
        & & Avg & Range & Avg & Range \\
        \midrule
        Level 1 (Simple) & 133 & 1.5 & 1--2 & 1.9 & 1--2 \\
        Level 2 (Medium) & 92  & 2.3 & 2--4 & 3.7 & 3--4 \\
        Level 3 (Hard)   & 51  & 4.4 & 3--7 & 5.7 & 3--7 \\
        \midrule
        \textbf{Overall} & \textbf{276} & \textbf{2.6} & \textbf{1--7} & \textbf{3.1} & \textbf{1--7} \\
        \bottomrule[1.5pt]
    \end{tabular}
\end{table*}

\subsection{Application Ecosystem and Information Spaces}
To ensure ecological validity and broad coverage, \Benchmark~integrates a diverse ecosystem of \textbf{25 applications}, prioritizing platforms that host Immutable Historical Records or Fixed Physical Attributes.
As detailed in Table \ref{tab:app_list}, the selection strictly adheres to the \textit{"Determinism Principle"}: apps like \textit{FlightAware} (Historical Flights) and \textit{Steam} (Software Specs) provide indisputable verification sources that are immune to temporal drift.

\begin{table*}[h]
\centering
\caption{The Application Ecosystem of \Benchmark}
\label{tab:app_list}
\small 
\begin{tabular}{l p{4.5cm} p{6.5cm}}
\toprule[1.5pt]
\textbf{Category} & \textbf{Applications} & \textbf{Primary Information Space (Sources of Ground Truth)} \\
\midrule
\textbf{Global Knowledge} & Wikipedia, Google Search & \textbf{Universal Facts:} Historical events, geographical data, official entity names. \\
\midrule
\textbf{Computational Knowledge} & WolframAlpha & \textbf{Scientific Facts:} Historical weather data, astronomical timestamps, chemical element properties, mathematical constants. \\
\midrule
\textbf{Multimedia (Audio)} & Spotify, Apple Music, NetEase Cloud & \textbf{Discography Data:} Album release dates, track duration, artist labels, explicit content tags. \\
\midrule
\textbf{Multimedia (Video)} & YouTube, Bilibili & \textbf{Video Metadata:} Upload timestamps, channel/UP-loader names, video duration, view count snapshots. \\
\midrule
\textbf{Geo-Spatial} & Google Maps, Amap (Gaode) & \textbf{POI \& Navigation:} Business addresses, subway exit numbers, historical traffic data, distance estimates. \\
\midrule
\textbf{Movies \& TV} & IMDb, Douban & \textbf{Filmography:} Cast lists, director credits, runtime, release years, award history. \\
\midrule
\textbf{Finance \& Commerce} & Yahoo Finance, Amazon & \textbf{Market Data:} Historical stock closing prices, product dimensions, ISBNs, publication dates. \\
\midrule
\textbf{Sports} & ESPN / NBA App & \textbf{Match Records:} Final scores, game dates, player statistics, MVP awards. \\
\midrule
\textbf{Travel \& Logistics} & Booking.com, TripAdvisor, FlightAware & \textbf{Itinerary Facts:} Hotel check-in policies, historical flight arrival times, gate records, historical establishment awards. \\
\midrule
\textbf{Digital Entertainment} & Steam, Epic Games & \textbf{Software Specs:} Release dates, developer names, minimum system requirements (OS/RAM), achievement lists. \\
\midrule
\textbf{System Utilities} & Messages, Contacts, Calendar, Photos, Calculator, Files, Clock & \textbf{Local Context:} Sender IDs, meeting slots, EXIF timestamps, file paths, alarm settings. \\
\bottomrule[1.5pt]
\end{tabular}
\end{table*}

\subsection{Task example of \Benchmark}
\label{append:Task example}
\begin{figure*}[ht]
    \centering
    \includegraphics[width=0.8\linewidth]{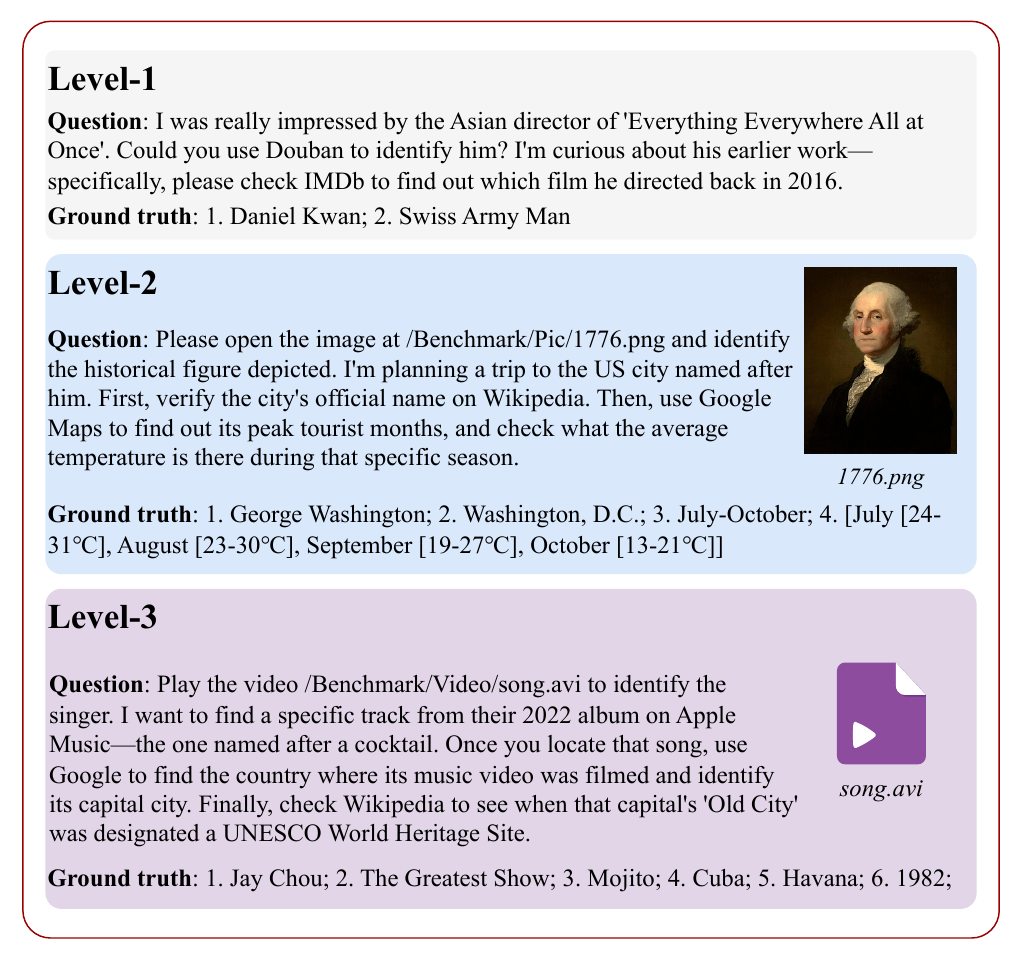}
    \caption{Task examples of different difficulty levels in \Benchmark}
    \label{fig:level example}
\end{figure*}

Figure~\ref{fig:level example} presents concrete examples of tasks corresponding to the three difficulty levels. These examples illustrate the increasing requirements in terms of the number of atomic tasks, the diversity of applications involved, and the complexity of planning and execution across the levels.

\section{Example of task execution trajectory}
\label{append:ET-Example}
As shown in the fig\ref{append:fig:et_2} and code\ref{lst:Example-of-ET2}, we demonstrate the complete execution trajectory of a GUI execution task that consists of 3 atomic tasks.

\section{Systematic Benchmark Comparison}
\label{append:benchmark_comparison}
To clearly demonstrate \Benchmark's unique niche, Table~\ref{tab:benchmark_comparison} provides a systematic comparison with current mainstream benchmarks from multiple dimensions. Compared to existing works like AndroidWorld and Mind2Web, \Benchmark's core advantage lies in providing fine-grained traversal metrics (MATCR, ATSR) to measure the agent's intermediate inference progress. Furthermore, by employing logico-linguistic decoupling and causal paths, it maintains an infinite number of dynamic data instances while ensuring strict state verifiability (combining device state with text match), effectively preventing data contamination.

\begin{table*}[h]
    \centering
    \small
    \resizebox{\textwidth}{!}{
    \begin{tabular}{llclcl}
        \toprule
        \textbf{Benchmark} & \textbf{Environment} & \textbf{Tasks} & \textbf{Verifiable State} & \textbf{Linguistic Decoupling} & \textbf{Evaluation Granularity} \\
        \midrule
        Mind2Web & Desktop Web & 2350 & None (Action Match) & No & Single-level \\
        OSWorld & Desktop/Web & 369 & Cloud/Device State & No & Single-level \\
        WebVoyager & Desktop Web & 643 & LLM Judge & No & Single-level \\
        AndroidWorld & Android & 116 & Device State & No & Single-level \\
        \rowcolor{gray!15} \Benchmark~(Ours) & Android/Desktop & 276 & Device State + Text Match & Yes (Causal Paths) & Multi-level (WPSR, MATCR, ATSR) \\
        \bottomrule
    \end{tabular}
    }
    \caption{Systematic comparison of \Benchmark~with existing GUI agent benchmarks.}
    \label{tab:benchmark_comparison}
\end{table*}

\subsection{Generalization on External Benchmarks (AndroidWorld)}
To dispel concerns regarding the framework's feasibility and generalizability, we deployed \Agent-\AndroidA~on AndroidWorld, a completely independent and dynamic verifiable benchmark. As shown in Table~\ref{tab:androidworld_supp}, under a unified benchmark, our system equipped with Qwen3.5-397b-a17b achieved a success rate of 63.3\%, significantly outperforming the standard MAE. This confirms that the core contribution of our system lies in the decoupling of macro-planning and micro-execution, rather than simply relying on the design of the underlying accessibility tree.

\begin{table}[h]
    \centering
    \small
    \resizebox{\linewidth}{!}{
    \begin{tabular}{lllc}
        \toprule
        \textbf{Agent} & \textbf{Model} & \textbf{Obs. Space} & \textbf{SR (P@1)} \\
        \midrule
        MAe & Qwen3.5-397b-a17b & Screenshot & 46.7 \\
        \Agent-MAe & Qwen3.5-397b-a17b & Screenshot & 50.0 \\
        \Agent-\AndroidA & Qwen3.5-397b-a17b & Scrn + A11y tree & 63.3 \\
        \midrule
        M3A & UI-TARS-1.5-7b & Scrn + A11y tree & 33.3 \\
        \Agent-\AndroidA & UI-TARS-1.5-7b & Scrn + A11y tree & 36.7 \\
        \bottomrule
    \end{tabular}
    }
    \caption{Cross-platform verification experiments on the AndroidWorld benchmark.}
    \label{tab:androidworld_supp}
\end{table}

\subsection{Comparative Analysis with Web-Specific Frameworks (Mind2Web)}
Mind2Web is primarily designed for static Web DOM tree parsing. To ensure a fair comparison, we extracted a subset of web tasks from \Benchmark~that can be completed using only a browser and tested them using the Mind2Web framework with the same Qwen3.5 model. As shown in Table~\ref{tab:mind2web_supp}, \Agent-\AndroidA~significantly outperforms Mind2Web (Overall SR 31.4\% vs 22.9\%). This performance gap stems from two fundamental differences: (1) Jarvis's hybrid vision-structure awareness effectively avoids element positioning failures caused by relying solely on dynamic, non-standard DOM trees; (2) The LightManus macro-planner mitigates semantic drift in long-sequence execution through Context Evolution, enabling it to maintain higher path completion quality in Level-2 and Level-3 tasks compared to Mind2Web's static memory.

\begin{table*}[h]
    \centering
    \small
    \setlength{\tabcolsep}{3pt}
    \resizebox{\textwidth}{!}{
    \begin{tabular}{l cccc cccc cccc cccc}
        \toprule
        \multirow{2}{*}{\textbf{Method}} & \multicolumn{4}{c}{\textbf{Level-1}} & \multicolumn{4}{c}{\textbf{Level-2}} & \multicolumn{4}{c}{\textbf{Level-3}} & \multicolumn{4}{c}{\textbf{Overall}} \\
        \cmidrule(lr){2-5} \cmidrule(lr){6-9} \cmidrule(lr){10-13} \cmidrule(lr){14-17}
        & \textbf{SR (P@1/4)} & \textbf{WPSR} & \textbf{MATCR} & \textbf{ATSR} & \textbf{SR (P@1/4)} & \textbf{WPSR} & \textbf{MATCR} & \textbf{ATSR} & \textbf{SR (P@1/4)} & \textbf{WPSR} & \textbf{MATCR} & \textbf{ATSR} & \textbf{SR (P@1/4)} & \textbf{WPSR} & \textbf{MATCR} & \textbf{ATSR} \\
        \midrule
        Mind2Web & 40.0 / 86.7 & 49.1 & 58.3 & 54.9 & 10.0 / 50.0 & 20.0 & 36.9 & 33.0 & 10.0 / 30.0 & 17.8 & 29.2 & 27.9 & 22.9 / 60.0 & 22.7 & 43.9 & 32.6 \\
        \Agent-\AndroidA & 46.7 / 86.7 & 56.2 & 61.7 & 59.8 & 30.0 / 70.0 & 32.5 & 40.0 & 38.0 & 10.0 / 40.0 & 20.3 & 27.5 & 26.8 & 31.4 / 68.6 & 28.8 & 45.7 & 34.0 \\
        \bottomrule
    \end{tabular}
    }
    \caption{Performance comparison between \Agent-\AndroidA~and Mind2Web on a pure web subset of \Benchmark.}
    \label{tab:mind2web_supp}
\end{table*}

\section{Error Type Examples}
\label{append:Error Type}
\subsection*{Planning \& Reasoning Errors (PRE)}

\paragraph{Original Task:} "I'm trying to settle a debate about a film score. I need to identify a specific movie where a certain musician served as the music director in 2019. I've totally forgotten his name, but I know he was the arranger for the title track of an album released on June 5, 2015. That album belongs to the same singer who performed 'Crush You into an Apple'.

Could you act as a detective? First, use Google Search to identify the singer of that song. I also need her debut year for my notes, so please look that up on Wikipedia. Then, trace back to her 2015 album, find that arranger's name via Wikipedia, and finally use Google Search to tell me which movie he directed the music for in 2019."

\noindent \textbf{Atomic tasks partitioned by LLM:}
\begin{enumerate}
    \item "Use Google Search to determine which album includes the song 'Crush You into an Apple', and identify the performer of that album from the search results snippet/page."
    \item "Open the Wikipedia app and search for the singer of "Alien from the Sky" to find the year they debuted."
    \item "Use Google Search to determine which album Joker Xue released on 2015-06-05."
    \item "Use Wikipedia to open the page for the album identified in task 3 as being released in 2005, and identify the arranger  of the album's title track (the song with the same name as the album)."
    \item "Use Google Search to search the arranger identified in task 4 and find which film they served as music director  for in 2019."
\end{enumerate}

\paragraph{Error analysis:} The Planning \& Reasoning Error (PRE) in this case stems from inconsistent task granularity during decomposition. The agent combined the dependent steps of "identifying the album" and "identifying the performer" into a single atomic task, causing a 1-step alignment shift for all subsequent IDs. 

\subsection*{Structural Compliance Errors(SCE)}

\textbf{The action given by LLM:} \\
\texttt{"action": "clear\_text(uid(5))"}

\paragraph{Error analysis:} "The returned action format is incorrect. It should be \texttt{clear\_text(5)}; the formatting error caused the action to fail."

\subsection*{Operational Errors(OE)}
 "The 'Expand All' button should have been clicked, but it clicked on Project Satan instead. The click position was incorrect."

\subsection*{Knowledge Deficits(KD)}

\paragraph{Error analysis:} The LLM lacks specific knowledge on how to operate Map applications. For the task of searching for the nearest major commercial district near Jiangnanxi Metro Station, Guangzhou, Guangdong Province, China, the model only queried the station name itself. Consequently, it failed to retrieve nearby commercial information. The correct search query should have been: major commercial districts near Jiangnanxi Metro Station, Guangzhou, Guangdong Province, China.

\subsection*{Perceptual Errors}
\paragraph{Task Description:} Use Google to find out which dynasty China was in during the year 1776.

\paragraph{Error Analysis:} The current screenshot already contains the required answer. However, the LLM misjudged the situation, and failed to extract the available information and instead returned to the home screen, leading to an unnecessary execution step.

\begin{figure*}[ht]
    \centering
    \includegraphics[width=0.5\linewidth]{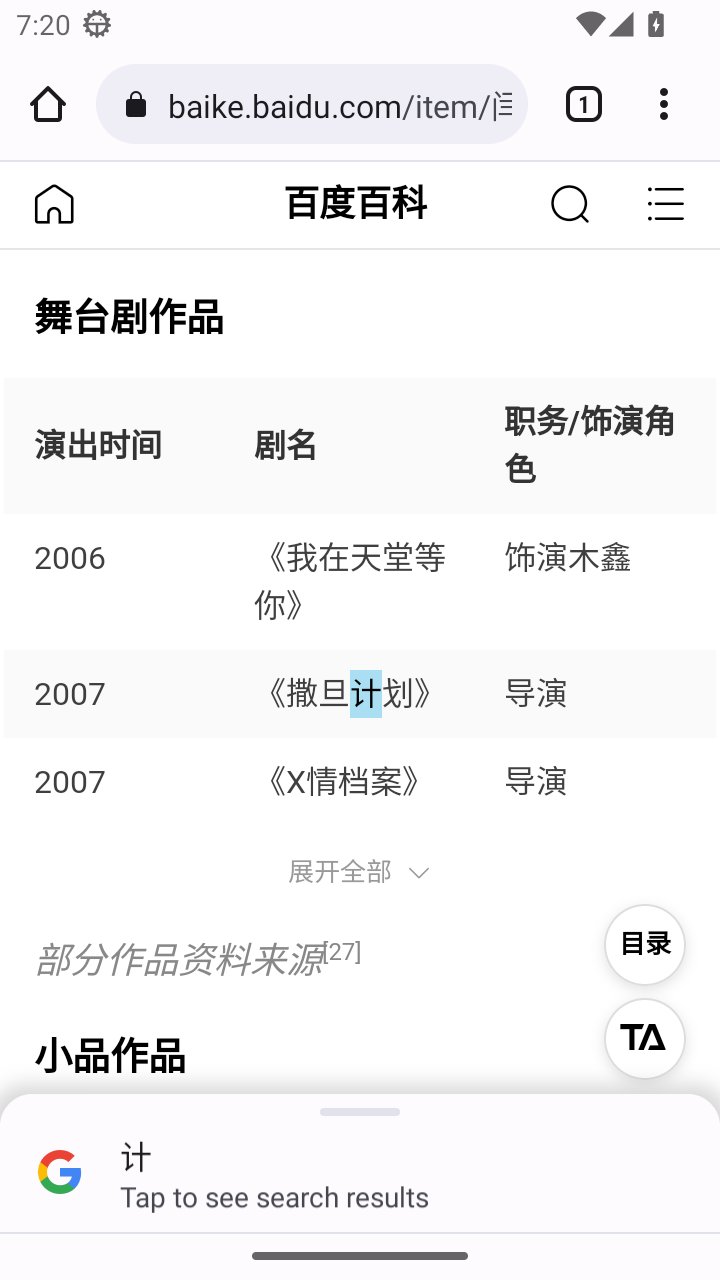}
    \caption{Operational Error Example}
    \label{append:fig:et_3}
\end{figure*}
\begin{figure*}[ht]
    \centering
    \includegraphics[width=0.5\linewidth]{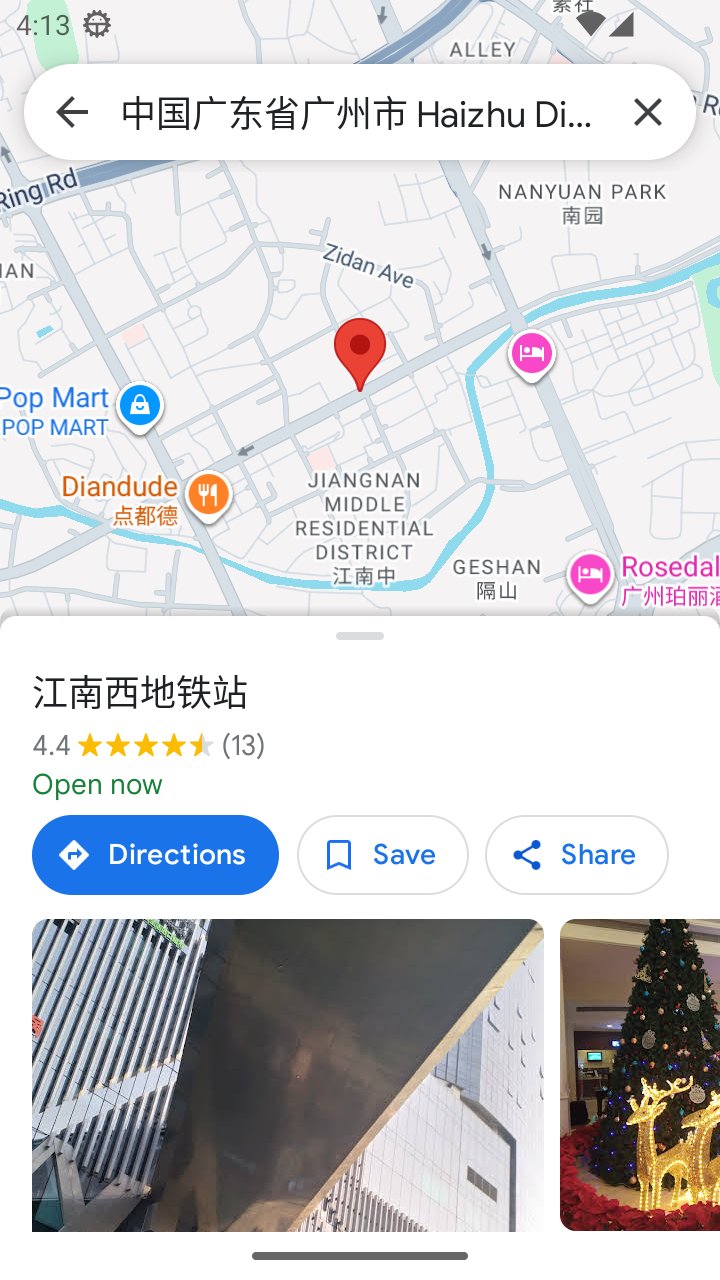}
    \caption{Operational Error Example}
    \label{append:fig:et_3}
\end{figure*}
\begin{figure*}[ht]
    \centering
    \includegraphics[width=0.5\linewidth]{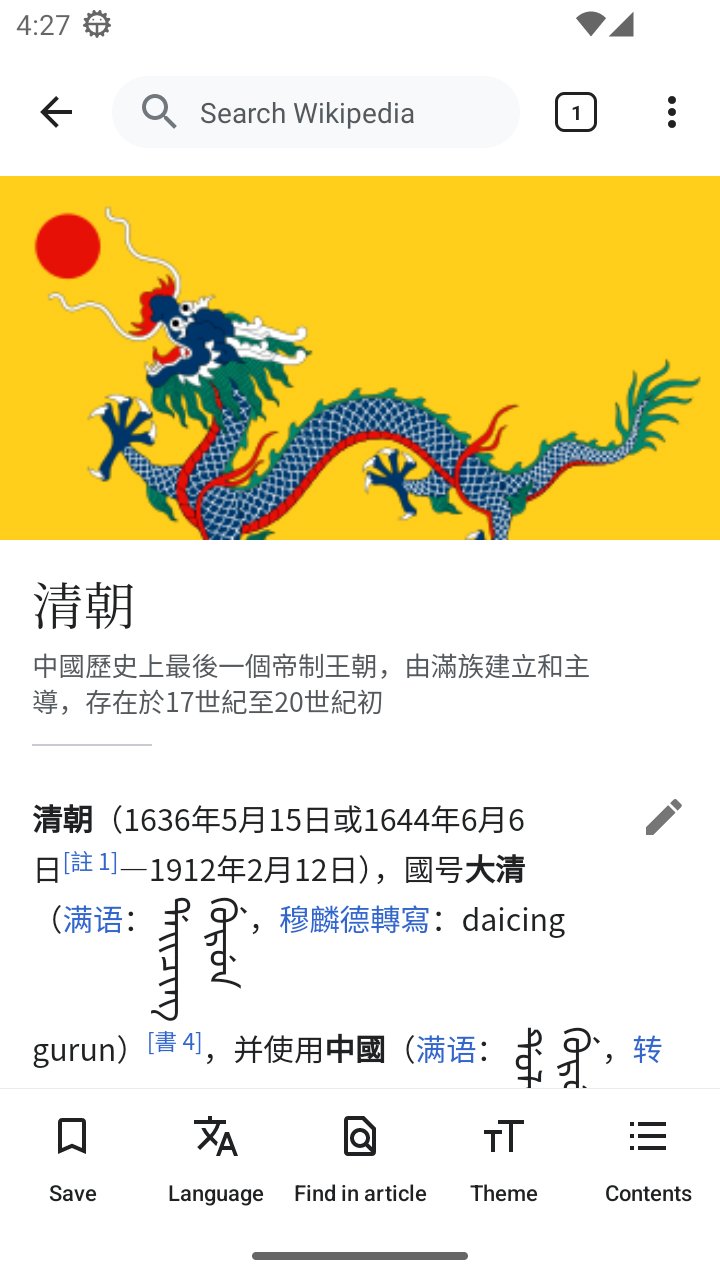}
    \caption{Operational Error Example}
    \label{append:fig:et_3}
\end{figure*}

\onecolumn
\definecolor{jsonBackground}{RGB}{250, 250, 250}   
\definecolor{jsonString}{RGB}{42, 161, 152}        
\definecolor{jsonKey}{RGB}{38, 139, 210}           
\definecolor{jsonNumber}{RGB}{211, 54, 130}        
\definecolor{customTag}{RGB}{133, 153, 0}          
\definecolor{customSection}{RGB}{203, 75, 22}      
\definecolor{customBold}{RGB}{0, 0, 0}             

\lstdefinestyle{json-custom}{
    backgroundcolor=\color{jsonBackground},
    basicstyle=\ttfamily\footnotesize,   
    breakatwhitespace=false,
    breaklines=true,                     
    captionpos=b,                        
    keepspaces=true,                     
    numbers=left,                        
    numbersep=10pt,                      
    numberstyle=\tiny\color{gray},       
    showspaces=false,
    showstringspaces=false,              
    showtabs=false,
    tabsize=2,                           
    frame=lines,                         
    rulecolor=\color{black!20},          
    %
    stringstyle=\color{jsonString},
    morestring=[b]",                     
    %
    moredelim=[s][\color{customSection}\bfseries]{===}{===},
    moredelim=[s][\color{customTag}]{<}{>},
    moredelim=[s][\color{black}\bfseries]{**}{**},
    %
    literate=
     *{0}{{{\color{jsonNumber}0}}}{1}
      {1}{{{\color{jsonNumber}1}}}{1}
      {2}{{{\color{jsonNumber}2}}}{1}
      {3}{{{\color{jsonNumber}3}}}{1}
      {4}{{{\color{jsonNumber}4}}}{1}
      {5}{{{\color{jsonNumber}5}}}{1}
      {6}{{{\color{jsonNumber}6}}}{1}
      {7}{{{\color{jsonNumber}7}}}{1}
      {8}{{{\color{jsonNumber}8}}}{1}
      {9}{{{\color{jsonNumber}9}}}{1}
      {:}{{{\color{black}:}}}{1}      
      {,}{{{\color{black},}}}{1}      
      {\{}{{{\color{black}\{}}}{1}    
      {\}}{{{\color{black}\}}}}{1}    
      {[}{{{\color{black}[}}}{1}      
      {]}{{{\color{black}]}}}{1},     
}

\begin{center}
    \includegraphics[width=0.7\linewidth]{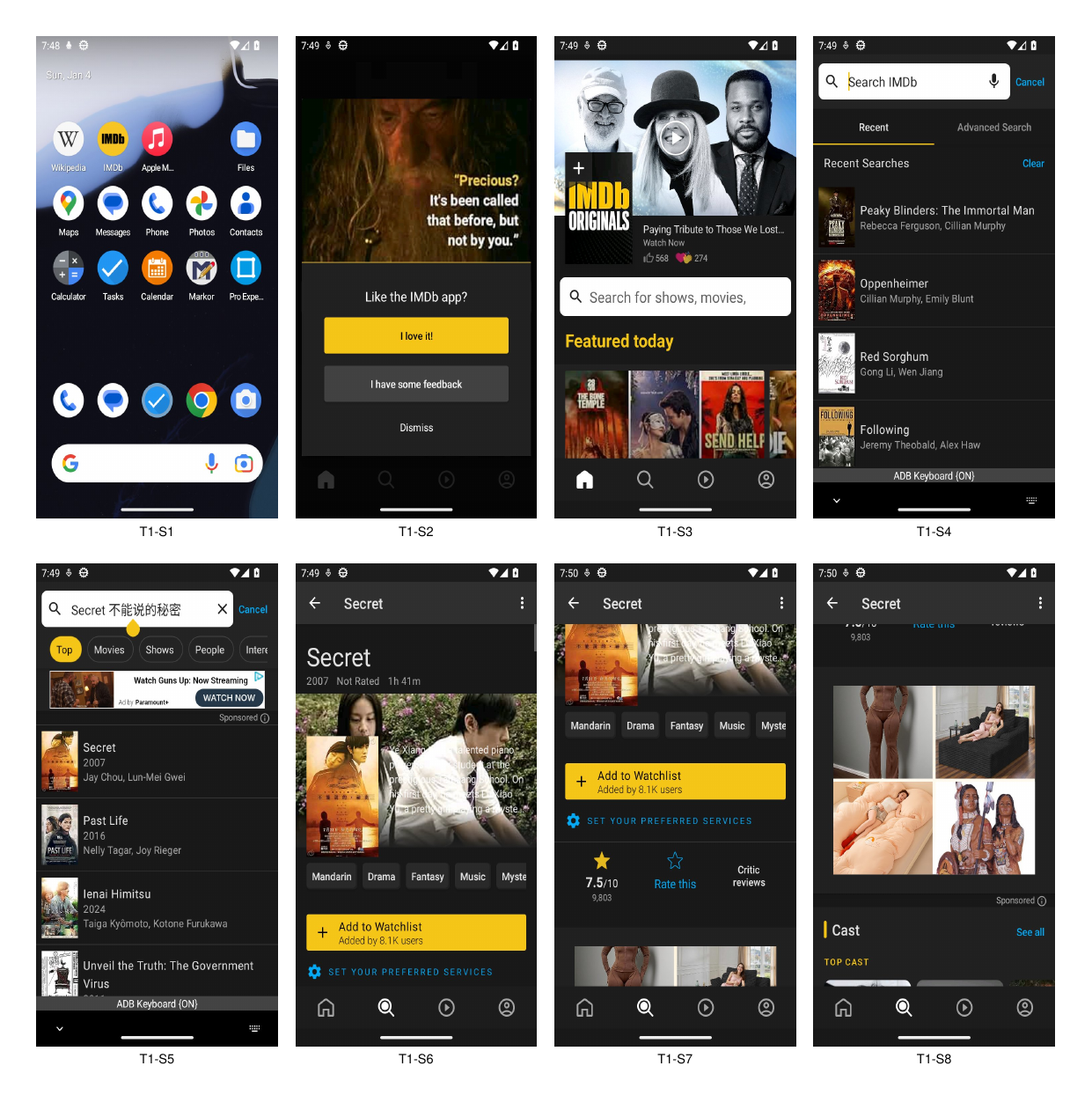}
    \includegraphics[width=0.7\linewidth]{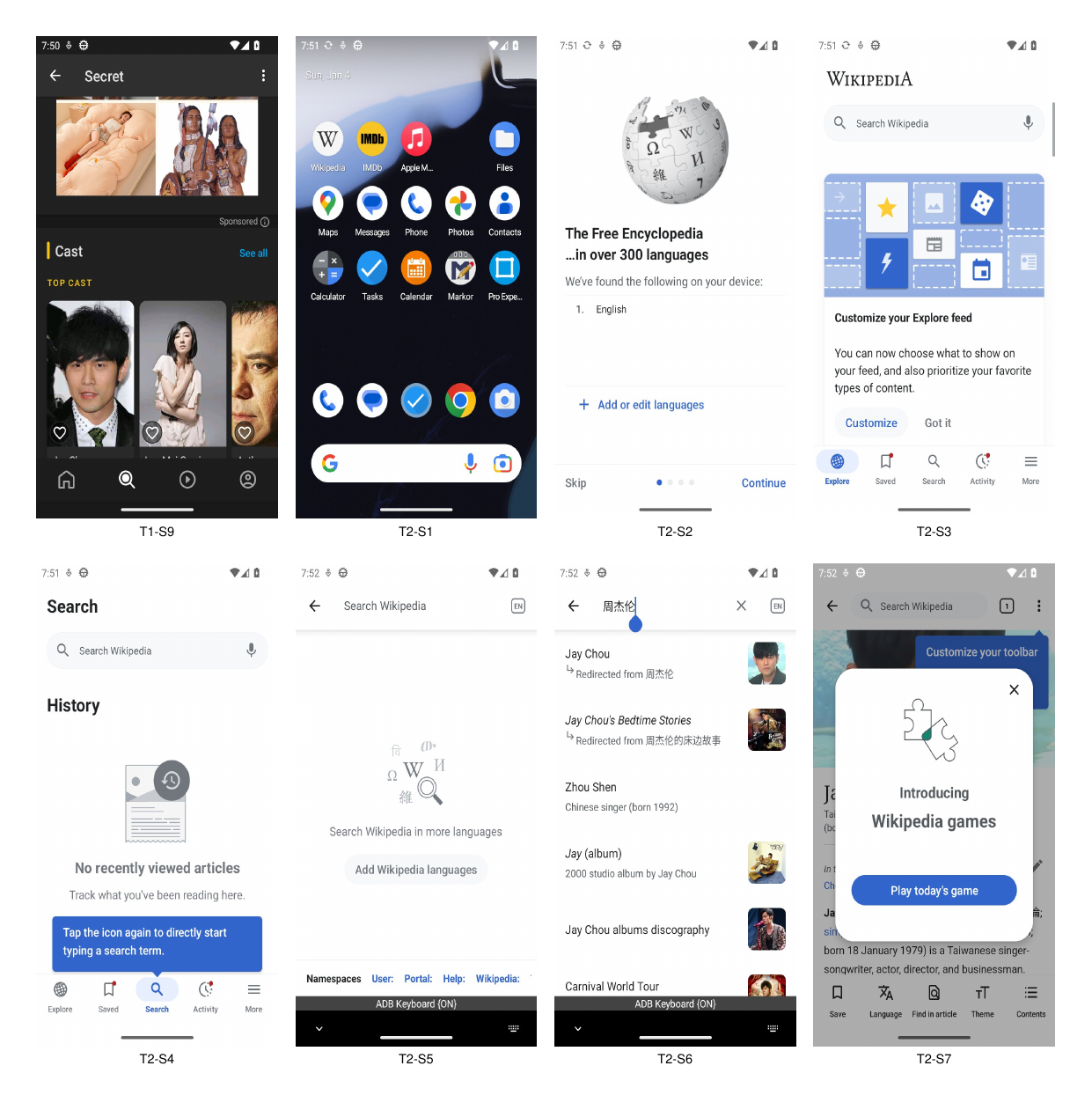}
    \includegraphics[width=0.7\linewidth]{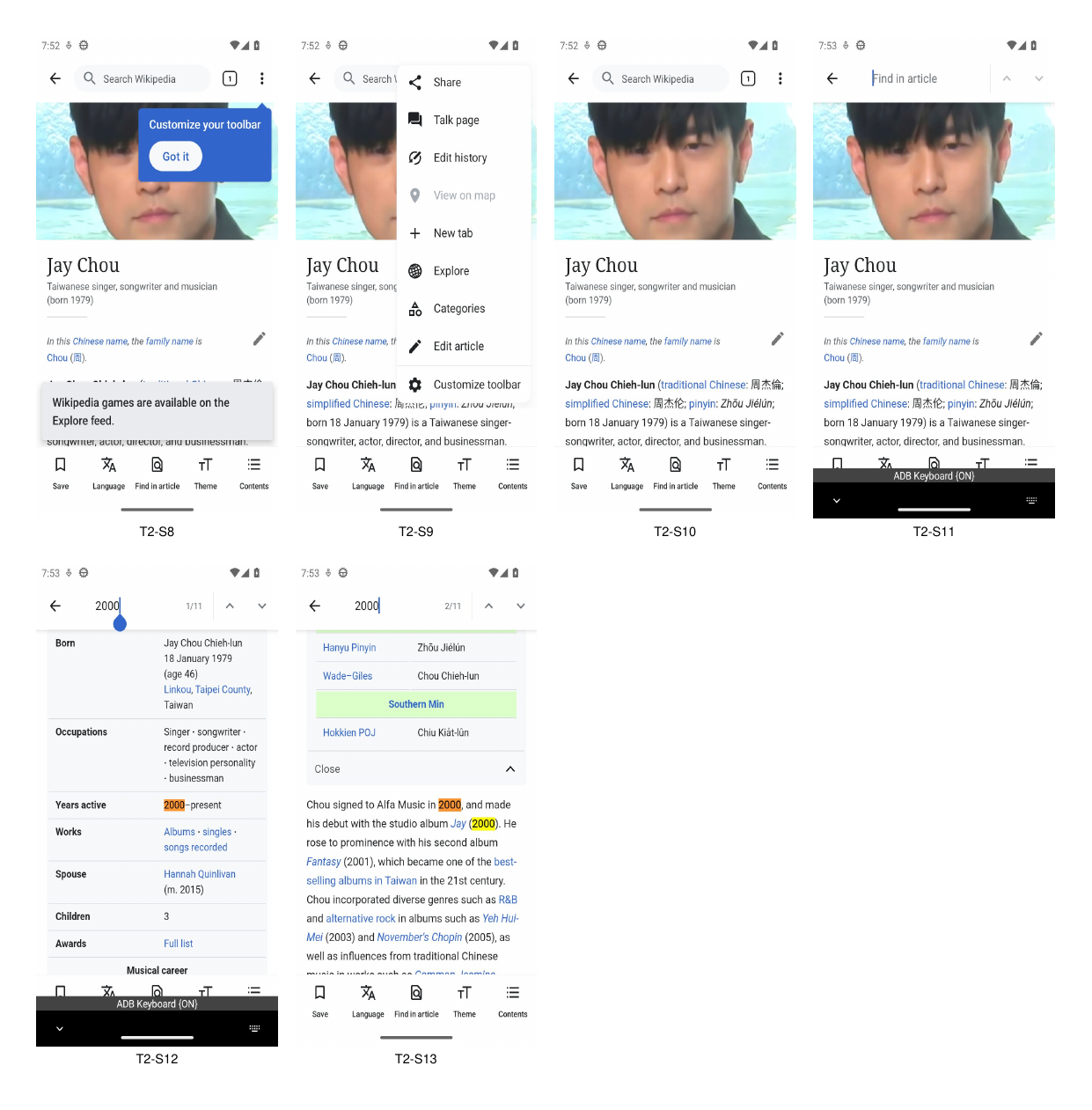}
    
    \captionof{figure}{Image from \texttt{image\_10-42\_1\_1.jpg} to \texttt{image\_10-42\_2\_13.jpg}}
    \label{append:fig:et_2}
\end{center}

\begin{lstlisting}[style=json-custom, caption={Example of Execution Trajectory}, label={lst:Example-of-ET2}]
{
    "images": [
        "image_-10-42_1_1.jpg",
        "image_-10-42_1_2.jpg",
        "image_-10-42_1_3.jpg",
        "image_-10-42_1_4.jpg",
        "image_-10-42_1_5.jpg",
        "image_-10-42_1_6.jpg",
        "image_-10-42_1_7.jpg",
        "image_-10-42_1_8.jpg",
        "image_-10-42_1_9.jpg",
        "image_-10-42_2_1.jpg",
        "image_-10-42_2_2.jpg",
        "image_-10-42_2_3.jpg",
        "image_-10-42_2_4.jpg",
        "image_-10-42_2_5.jpg",
        "image_-10-42_2_6.jpg",
        "image_-10-42_2_7.jpg",
        "image_-10-42_2_8.jpg",
        "image_-10-42_2_9.jpg",
        "image_-10-42_2_10.jpg",
        "image_-10-42_2_11.jpg",
        "image_-10-42_2_12.jpg",
        "image_-10-42_2_13.jpg"
    ],
    "original_task": [

    ]

===Task1===

    "overall_task": [
	    "Use IMDB to search for the movie 'Secret' and identify the male lead actor."
    ]

    ===Step1===

    <image_-10-41_1_1.jpg>

    **Thought**
    We need to use the IMDb app to search for the movie, so first open IMDb from the home screen.

    **Action**
    tap(5)

    ===Step2===

    <image_-10-41_1_2.jpg>

    **Thought**
    IMDb opened successfully but a feedback popup is blocking the app. I need to dismiss it to access search.

    **Action**
    tap(4)

    ===Step3===

    <image_-10-41_1_3.jpg>

    **Thought**
    The feedback popup is gone and the IMDb home screen is visible with a prominent search field. Next step is to open the search input to look up the movie title.

    **Action**
    tap(13)

    ===Step4===

    <image_-10-41_1_4.jpg>

    **Thought**
    Search screen is open with an editable field labeled 'Search IMDb'. To find the movie 'Secret', I should type the title into this field and submit.

    **Action**
    input_text(3, "Secret")

    ===Step5===

    <image_-10-41_1_5.jpg>

    **Thought**
    The query is entered and results are shown. The top result is the movie 'Secret' (2007) with cast listed. Open this movie's page to confirm the male lead actor.

    **Action**
    tap(31)

    ===Step6===

    <image_-10-41_1_6.jpg>

    **Thought**
    Now on the IMDb title page for 'Secret' (2007). To identify the male lead actor, I need to find the cast/credits section on this page, which likely requires scrolling down to the 'Top cast' area.

    **Action**
    swipe("UP","MEDIUM")

    ===Step7===

    <image_-10-41_1_7.jpg>

    **Thought**
    The swipe did not reveal the cast section yet; we're still near the top with plot, genres, and rating. To find the male lead actor, I need to scroll further down until the 'Top cast' appears.

    **Action**
    swipe("UP","LONG")

    ===Step8===

    <image_-10-41_1_8.jpg>

    **Thought**
    The swipe was successful: the Cast section with TOP CAST is now visible. To identify the male lead actor, I need to view the top cast names; they should be in the visible list below, so I will scroll slightly to reveal the first cast entries.

    **Action**
    swipe("UP","MEDIUM")

    ===Step9===

    <image_-10-41_1_9.jpg>

    **Thought**
    The swipe revealed the Top Cast list with actor names. The first listed male actor is Jay Chou, which indicates the male lead. The task is to identify the male lead actor, so I can complete now.

    **Action**
    finish("On IMDb for 'Secret', the male lead actor listed in Top Cast is Jay Chou.")

===Task2===

    "overall_task": [
	    "Use Wikipedia to search for Jay Chou and find which album he released in 2000."
    ]

    ===Step1===

    <image_-10-41_2_1.jpg>

    **Thought**
    We need to use Wikipedia to search for Jay Chou, so first open the Wikipedia app from the home screen.

    **Action**
    tap(4)

    ===Step2===

    <image_-10-41_2_2.jpg>

    **Thought**
    Wikipedia app opened to an onboarding language selection screen that blocks access to search. To proceed quickly to the app and search for Jay Chou, skip onboarding.

    **Action**
    tap(8)

    ===Step3===

    <image_-10-41_2_3.jpg>

    **Thought**
    Onboarding was skipped successfully; we are now on Wikipedia Explore with a banner. To search for Jay Chou, open the Search tab.

    **Action**
    tap(15)

    ===Step4===

    <image_-10-41_2_4.jpg>

    **Thought**
    We're on Wikipedia's Search tab with a visible 'Search Wikipedia' field. Next step is to tap the search field and type Jay Chou to find his page and then identify the 2000 album.

    **Action**
    tap(4)

    ===Step5===

    <image_-10-41_2_5.jpg>

    **Thought**
    The tap opened the Wikipedia search interface with the editable search text field visible. Next we need to enter Jay Chou to pull up the correct article.

    **Action**
    input_text(4, "Jay Chou")

    ===Step6===

    <image_-10-41_2_6.jpg>

    **Thought**
    Search results for Jay Chou are displayed. To find which album he released in 2000, open the main Jay Chou article result.

    **Action**
    tap(10)

    ===Step7===

    <image_-10-41_2_7.jpg>

    **Thought**
    A popup about Wikipedia games is blocking the article content, so I need to close it before reading the Jay Chou page to find the 2000 album.

    **Action**
    tap(5)

    ===Step8===

    <image_-10-41_2_8.jpg>

    **Thought**
    A modal popup titled "Introducing Wikipedia games" is still blocking the Jay Chou article; I need to close it using the X in the popup to access the content and find the 2000 album.

    **Action**
    tap(12)

    ===Step9===

    <image_-10-41_2_9.jpg>

    **Thought**
    The popups are gone, but a three-dot overflow menu is open covering the article. I need to dismiss this menu to access the article content and find the 2000 album.

    **Action**
    back()

    ===Step10===

    <image_-10-41_2_10.jpg>

    **Thought**
    The overflow menu is now dismissed and I can see the Jay Chou article. To find the album released in 2000 quickly, I should use the built-in "Find in article" to search for 2000 or "studio albums."

    **Action**
    tap(46)

    ===Step11===

    <image_-10-41_2_11.jpg>

    **Thought**
    The Find in article bar is open with an input field. To locate the relevant section about albums released in 2000, I should search within the article for the year 2000.

    **Action**
    input_text(56, "2000")

    ===Step12===

    <image_-10-41_2_12.jpg>

    **Thought**
    The in-page find for "2000" is active and shows 1/11 matches; we're currently on the infobox match (Years active 2000-present), not the album release. I should jump to the next match to find where the 2000 album is mentioned.

    **Action**
    tap(89)

    ===Step13===

    <image_-10-41_2_13.jpg>

    **Thought**
    The next match (2/11) highlights text stating he made his debut with the studio album 'Jay' (2000), which answers which album he released in 2000. I should end the find mode and report the album.

    **Action**
    finish("On Jay Chou's Wikipedia page, the text says he made his debut with the studio album 'Jay' (2000). So the album released in 2000 is: Jay.")
}

\end{lstlisting}

\end{document}